\documentclass[twoside]{article}

\usepackage[accepted]{aistats2023}
\usepackage[round]{natbib}

\usepackage{url}
\usepackage{xr-hyper}
\usepackage{hyperref}
\usepackage{macros}
\usepackage[capitalize,noabbrev]{cleveref}
\usepackage{xcolor}
\usepackage{booktabs}
\usepackage{amsfonts}
\usepackage{nicefrac}
\usepackage{microtype}
\usepackage{caption}
\usepackage{subfig}
\usepackage{multirow}
\usepackage{paralist}
\usepackage[T1]{fontenc}

\usepackage{algorithm2e}
\RestyleAlgo{ruled}
\crefname{algocf}{algorithm}{algorithms}
\Crefname{algocf}{Algorithm}{Algorithms}
\usepackage{color}

\begin{document}

%

%
\runningauthor{T. Cinquin, T. Rukkat, P. Schmidt, M. Wistuba, A. Bekasov}

\twocolumn[

\aistatstitle{Variational Boosted Soft Trees}

\aistatsauthor{
  Tristan Cinquin \\ University of Tuebingen\footnotemark[1] \\ \href{mailto:tristan.cinquin@uni-tuebingen.de}{\texttt{tristan.cinquin@uni-tuebingen.de}}
  \And Tammo Rukat \\ Deekard\footnotemark[1] \\ \href{mailto:tammorukat@gmail.com}{\texttt{tammorukat@gmail.com}}
  \AND Philipp Schmidt \\ Amazon \\ \href{mailto:phschmid@amazon.com}{\texttt{phschmid@amazon.com}}
  \And Martin Wistuba \\ Amazon \\ \href{mailto:marwistu@amazon.com}{\texttt{marwistu@amazon.com}}
  \And Artur Bekasov \\ Amazon \\ \href{mailto:abksv@amazon.com}{\texttt{abksv@amazon.com}}
}

\aistatsaddress{} ]

\begin{abstract}
Gradient boosting machines (GBMs) based on decision trees consistently demonstrate state-of-the-art results on regression and classification tasks with tabular data, often outperforming deep neural networks.
However, these models do not provide well-calibrated predictive uncertainties, which prevents their use for decision making in high-risk applications. 
The Bayesian treatment is known to improve predictive uncertainty calibration, but previously proposed Bayesian GBM methods are either computationally expensive, or resort to crude approximations.
Variational inference is often used to implement Bayesian neural networks, but is difficult to apply to GBMs, because the decision trees used as weak learners are non-differentiable.
In this paper, we propose to implement Bayesian GBMs using variational inference with \emph{soft} decision trees, a fully differentiable alternative to standard decision trees introduced by \citeauthor{isroy2012softdecisiontree}
Our experiments demonstrate that variational soft trees and variational soft GBMs provide useful uncertainty estimates, while retaining good predictive performance.
The proposed models show higher test likelihoods when compared to the state-of-the-art Bayesian GBMs in 7/10 tabular regression datasets and improved out-of-distribution detection in 5/10 datasets.
\end{abstract}

\footnotetext[1]{Work done while at Amazon.}

\section{INTRODUCTION}

Tabular data, often combining both real valued and categorical variables, is common in machine learning applications \citep{borisov2021DLTDsurvey, Hoang2021MachineLM, clements2020DLcreditmonitoring, Ebbehoj2021TLforHCML, computers10020024}.
A number of these applications require uncertainty estimates, in addition to the model's predictions.
Predictive uncertainty is particularly important when using the model for making decisions associated with risk, as common in healthcare \citep{medicineBDL, uncertaintymedicalML} or finance \citep{analystRecommendationBML, bayesMLinFinance}. 
In order to make optimal decisions, we require the uncertainty estimates to be \emph{well-calibrated}: the predictive distribution returned by the model should capture the true likelihood of the targets \citep{guo2017calibration}.

A \emph{decision tree} is a 
predictive model that is fit by recursively partitioning the feature space.
These models are fast, interpretable and support categorical data without additional pre-processing.
To increase their expressivity without exacerbating overfitting, multiple decision trees can be combined into an ensemble using \emph{gradient boosting} \citep{FREUND1997Adaboost, Friedman2001GBM}.
\emph{Gradient boosting machines} (GBM) are an expressive class of machine learning models trained by sequentially fitting an ensemble of weak learners (\eg decision trees). 
At each iteration, the gradient boosting procedure fits an additional weak learner that would minimize the training error when added to the ensemble.
These models demonstrate state-of-the-art results on tabular data \citep{Bojer_2021, liuhal02957135}, often out-performing deep neural networks \citep{borisov2021DLTDsurvey, SHWARTZZIV202284}.

While GBM models demonstrate good predictive performance, they do not provide well-calibrated predictive uncertainties \citep{Niculescu2012calibrationGBM}.
Bayesian methods are known to improve predictive uncertainties by explicitly capturing the \emph{epistemic} uncertainty, a type of uncertainty that results from learning with finite data \citep{kristiadi2020bayesian, mitros2019validity}.
Bayesian GBMs have been proposed, and indeed improve upon the predictive uncertainties of standard GBMs \citep{chipman1998BayesianCART, linero2017SBART, ustimenko2020sglb}.
Current methods rely on \emph{Markov chain Monte Carlo} (MCMC) to sample from the posterior, however, and do not scale well to larger models and datasets.
Alternative approximate sampling methods have been proposed by \citet{he2020stochastictreeensemble} and \citet{malinin2021uncertGBM}.
These methods make the sampling procedure more efficient, but degrade the quality of the resulting predictive uncertainties.
More scalable approximate inference methods such as the Laplace approximation or variational inference are difficult to apply to GBMs, because the decision trees used as weak learners are non-differentiable. 
Unlike standard decision trees that route the input to a unique leaf based on a series of binary conditions, \emph{soft decision trees} \citep{isroy2012softdecisiontree} compute a convex combination of \emph{all} leaves, weighing their values by a series of learned gating functions.
As a result, soft decision trees are fully differentiable, and therefore amenable to scalable Bayesian inference methods that are common in Bayesian deep learning \citep{blundell2016mfvibnn, cobb2020hmc, immer2020BNNlocalLaplace}.

In this work we propose to implement Bayesian GBMs by using soft trees as week learners in a GBM, and performing variational inference on the resulting model. 
In particular, we make the following contributions: 
\begin{enumerate}
    \item We propose a method for performing variational inference in soft decision trees. The chosen variational distribution allows trading-off memory and computation for a richer posterior approximation. 
    \item We increase the expressivity of the model by using variational soft trees as weak learners in a GBM.
    \item We run experiments to demonstrate that the proposed models perform well on tabular data, yielding useful predictive uncertainties in regression, out-of-distribution detection and contextual bandits.
\end{enumerate}

\section{BACKGROUND}

\subsection{Bayesian inference}
\label{sec:bi_vi}

Bayesian inference provides a theoretical framework for reasoning about model uncertainty, and has been shown to improve calibration of machine learning models in practice \citep{mitros2019validity, kristiadi2020bayesian}. 
In Bayesian inference we place a prior on the model parameters~$\theta$, and use the Bayes rule to define the posterior distribution having observed the data:
\begin{align}
    \prob{\theta \g \setvar{D}} \propto \prob{\setvar{D} \g \theta} \prob{\theta}
\end{align}
where $\mathcal{D} = \set{x_i, y_i}_{i=1}^n$ is the training data, $\prob{\theta}$ is the prior distribution, and $\prob{\mathcal{D} \g \theta}$ the likelihood. The predictive distribution of a new target $y^*$ given a feature vector $x^*$ is then computed as
\begin{align}
    \prob{y^* \g x^*} = \expect{\prob{y^* \g x^*, \theta}}{\theta \sim \prob{\theta \g \setvar{D}}} \label{eq:bayesian_prediction}
\end{align}
The posterior distribution in the Bayesian paradigm
explicitly captures \emph{epistemic} or \emph{model} uncertainty that results from learning with finite data.
Epistemic uncertainty is one of the two fundamental types of uncertainty (alongside the \emph{aleatoric} or \emph{data} uncertainty), and capturing it was shown to be especially important
in risk-aware decision making \citep{riquelme2018DeepBayesianBandit, medicineBDL, deisenroth2011PILCO}, active learning \citep{kirsch2019BatchBald} and Bayesian optimisation \citep{bayesMLinFinance, krause2006sensorplacementGP, springenberg2016BOBNN}. 
Unfortunately, the posterior distribution is typically intractable and requires numerical approximation.


\paragraph{Variational inference}

Variational inference (VI) is an approximate inference method which selects a distribution $q^*$ within a variational family $\setvar{Q}$ that best approximates the posterior $p$.
More formally, we select the distribution $q^*$ such that:
\begin{equation}
\begin{aligned}
    q^* &= \argmin_{q \in \setvar{Q}} \kl{q}{p} \\ 
        &= \argmin_{q \in \setvar{Q}} \expect{\log \frac{q\br{\theta}}{\prob{\theta \g \setvar{D}}}}{\theta \sim q}.
\end{aligned}
\end{equation}
In practice, we define a parametric variational distribution $q_\phi$ and use standard optimization methods like stochastic gradient descent to minimize the KL loss \wrt $\phi$.
We then use $q^*\br{\theta}$ instead of $\prob{\theta \g \setvar{D}}$ to approximate the predictive distribution in \cref{eq:bayesian_prediction}. 
In this paper, we focus on VI because it can capture arbitrarily complex, high-dimensional posteriors, assuming the chosen variational family is sufficiently expressive.
At the same time, VI can scale to large datasets when combined with stochastic optimisation \citep{hoffman2013StochasticVI, swiatkowski2020ktied, tomczak2020lrvinn, tran2015vigp}.

Other commonly used approximate inference methods are Markov chain Monte Carlo (MCMC) \citep{welling2011sgld} and the Laplace approximation \citep{immer2020BNNlocalLaplace}. 
MCMC samples from the posterior directly, but does not scale well to large models and datasets. 
The Laplace approximation is more scalable, but assumes that the posterior can be approximated well with a Gaussian distribution. 

\subsection{Decision trees}

Decision trees are predictive models trained by recursively partitioning the input space and fitting a simple model (often constant) in each partition (leaf). 
At each non-terminal node of the tree, the model forwards the input to a child based on a binary condition.
For example, at a particular node the model might route the input $x$ to the left sub-tree if $x[i] \leq \tau$, and to the right sub-tree otherwise, where $i$ indexes a particular feature and $\tau$ is a threshold.
The splits are determined by greedily optimizing a chosen criterion at each node. 
For instance, in the regression setting the feature index $i$ and threshold $\tau$ could be selected to minimize the \emph{sum of square errors} of the induced partition.
The splitting hyperplanes are always parallel to the axes of the feature space. Predictions are made by forwarding $x$ to a unique leaf and outputting the class (for classification) or the real value (for regression) associated with this particular leaf.

\paragraph{Soft decision trees}

First introduced by \citet{isroy2012softdecisiontree}, soft decision trees use sigmoid gating functions to ``soften'' the binary routing at each node. 
More formally, the output $f\br{x}$ of a soft decision tree is computed as the sum of the leaf outputs $f_l\br{x}$ weighted by the probability $p_l\br{x}$ that the input $x$ is routed to the leaf $l$:
\begin{equation}
    f\br{x} = \sum_{l \in L} p_l\br{x} f_l\br{x},
\end{equation}
where $L$ is the set of leaf nodes.
The probability $p_l\br{x}$ is defined as the product of probabilities returned by the \emph{soft gating functions} on the path from the root to the leaf $l$.
Details on the exact form of $p_l$ and $f_l$ are provided in \cref{sec:vist_likelihood}.
The sigmoid gating functions allow learning non-axis-aligned decision boundaries and smooth functions.
This results in a differentiable model that can be trained by gradient descent, and yet performs well on tabular data \citep{feng2020sgbm, isroy2012softdecisiontree, linero2017SBART, Luo2021sdtr}. 
Computing the prediction of a soft tree requires evaluating \emph{all} the leaves, and is therefore more expensive than doing so for the standard decision tree.
In practice, however, a shallow soft decision tree can match the performance of a deep hard decision tree, which alleviates some of the computational overhead \citep{isroy2012softdecisiontree}.

Multiple variations of a soft decision tree exist. 
The tree depth can be fixed \citep{feng2020sgbm, kontschieder2015deepneuralforest, Luo2021sdtr} or adjusted dynamically by greedily adding new nodes while the validation loss decreases \citep{isroy2012softdecisiontree}. 
Furthermore, the sigmoid gating functions can be based on linear models \citep{feng2020sgbm, frosst1017distilling, isroy2012softdecisiontree, Luo2021sdtr}, multi-layer perceptrons \citep{feng2020sgbm, kontschieder2015deepneuralforest}, or convolutional neural networks \citep{Ahmetoglu2018csdt}. 
Likewise, the leaves of the tree can output constant values \citep{feng2020sgbm, frosst1017distilling, kontschieder2015deepneuralforest, Luo2021sdtr}, or be defined using complex models such as neural networks \citep{Ahmetoglu2018csdt}. 
To address the exponential increase in prediction cost as the soft tree grows, \citet{frosst1017distilling} propose to only consider the \emph{most probable} path from the root to a leaf. 
This method requires special regularization encouraging nodes to make equal use of each sub-tree to work and performs worse than averaging over all leaves, but is more efficient. 

In addition to state-of-the-art performance on tabular data, soft decision trees can be preferred to neural networks due to their greater interpretability \citep{frosst1017distilling}.
As soft trees rely on hierarchical decisions, one can examine the weights of the soft gating functions and gain insight into the learned function.
In prior work, neural networks have been distilled into into soft trees to make their decisions explainable, both in the context of image classification \citep{frosst1017distilling} and reinforcement learning \citep{Coppens2019DistillingDR}.

Soft decision trees are similar to \emph{hierarchical mixtures of experts} \citep[HME,][]{jordan1993HME}. While HMEs and soft decision trees share the same architecture, HMEs explicitly model the assignment of each input to a leaf using latent variables. 
Therefore, HMEs are trained using the expectation maximisation algorithm \citep[EM,][]{Dempster77em}, rather than by maximising the log-likelihood. 
\citeauthor{jordan1993HME} find that, when compared to the standard maximum likelihood with backpropagation, fitting an HME using EM requires fewer passes over the data, but the resulting models demonstrate higher test error.

\subsection{Gradient boosting machines}

A \emph{gradient boosting machine} (GBM) \citep{Friedman2001GBM} sequentially combines a set of weak learners (\eg decision trees) into an ensemble to obtain a stronger learner. 
At each iteration, a weak learner is trained to minimize the error of the existing ensemble, and is subsequently added to the ensemble.
More formally, at iteration $t$ the GBM combines the existing ensemble $f_t\br{x} = \sum_{j=1}^t \gamma_{j} h_{j}\br{x}$ with an additional learner $h_{t+1} \in \setvar{H}$ from some function class $\setvar{H}$ \st $f_{t+1}\br{x} = f_t\br{x} + \gamma_{t+1} h_{t+1}\br{x}$ minimizes the training loss.
Finding the optimal weak learner $h_{t+1}$ is intractable in general, so the optimization problem is simplified by taking a step in the direction of the steepest decrease of the loss $\mathcal{L}$ in function space:
\begin{equation}
    f_{t+1}\br{x} = f_t(x) - \gamma_{t+1} \sum_{i=1}^n\nabla_{f_t}\mathcal{L}(f_t\br{x_i}, y_i),
\end{equation}
where $\mathcal{D} = \set{x_i, y_i}_{i=1}^n$ is the training data. 
In other words, each weak learner $h_{t+1}$ is fit to match $-\gamma_{t+1} \sum_{i=1}^n\nabla_{f_t}\mathcal{L}(f_t\br{x_i}, y_i)$. 
Taking multiple gradient boosting steps minimizes the training loss and yields an accurate model. 
GBMs demonstrate state-of-the-art performance on tabular data for both classification and regression tasks: models that win Kaggle\footnote{\url{https://www.kaggle.com}} competitions are commonly based on GBMs \citep{Bojer_2021, liuhal02957135}. 
\section{RELATED WORK}

\paragraph{Bayesian hierarchical mixture of experts}


Prior work performed Bayesian inference on the parameters of HMEs using variational inference.
These methods consider an approximate posterior which factorizes the parameters into groups (leaf function, node gating function, latent variables, prior parameters) and places isotropic Gaussian priors on the leaf and node model parameters, as well as Gamma hyper-priors.
\citet{waterhouse1995BayesianHME} perform inference by iteratively optimizing the VI objective with respect to each group of parameters separately, keeping the others fixed. However, for tractability they use the Laplace approximation on the node gating function parameters and hence do not get a lower bound on the evidence.
\citet{ueda2002BayesianHME} propose to model the joint distribution over both input and output variables, which allows to derive the variational posterior without approximations for a tree of height 1. 
However, capturing the input distribution is both expensive and unnecessary for regression and classification. 
Also this method does not apply to deeper trees. 
Finally, \citet{bishop2002BayesianHME} derive a tractable lower bound on the variational inference objective, which is then optimised by iteratively updating the parameters of each factor of the variational posterior, keeping the others fixed. 

In contrast, we consider a simpler model which does not explicitly capture the assignment of data samples to leaves with latent variables. 
Furthermore, our proposed variational posterior is more expressive and can capture correlations between all the parameters of the model. 
Finally, we fit our model to optimise the variational objective using the reparameterization trick from \citet{kingma2013VAE} and stochastic gradient descent. 
\vspace{-0.2em}

\paragraph{Bayesian gradient boosted trees}

Previous work on Bayesian GBMs uses MCMC to sample from the Bayesian posterior.
\citet{chipman1998BayesianCART} suggest performing inference in a GBM using a Gibbs sampler which iteratively samples each tree conditioned on the other trees in the ensemble. 
This method, \emph{BART}, requires a long burn-in period (${\approx}1000$ iterations). 
\citet{pratola2017hbart} improve BART with a heteroscedastic noise model at the cost of making inference slower, as it is performed on both the mean and noise models. 
In order to speed up the convergence of the BART sampler, \citet{he2020stochastictreeensemble} propose \emph{Accelerated BART} (XBART). Rather than making small changes to a given tree at each iteration, XBART grows an entirely new tree. This speeds up mixing considerably requiring only 15 burn-in iterations. As a mechanism for variable selection, the authors also propose placing a Dirichlet prior on the input features, improving performance on high-dimensional data.

In parallel, \citet{ustimenko2020sglb} draw analogies between gradient boosting and MCMC, and prove that a set of modifications is enough to cast gradient boosting with Gaussian noise as stochastic gradient Langevin dynamics in function space. 
The resulting algorithm, \emph{stochastic gradient Langevin boosting} (SGLB), allows to draw multiple samples from the posterior by training an ensemble of GBMs with targets perturbed by Gaussian noise.
To reduce the memory and time complexity of SGLB, \citet{malinin2021uncertGBM} introduce \emph{virtual ensembles}. 
Taking advantage of the additive structure of GBMs, virtual ensembles collect multiple sub-ensembles by running a single SGLB model. 
However, faster training comes at the cost of higher correlation among models, thus reducing the effective sample size.

\paragraph{Soft gradient boosting machine}

Just like decision trees, soft decision trees can be combined into ensembles to improve their predictive performance.
Gradient boosting is a particularly successful method for doing so, hence \citet{Luo2021sdtr} experiment with combining soft trees using gradient boosting, and find the resulting method to be highly effective on tabular data. 
Similarly, \citet{feng2020sgbm} suggest to fit soft GBMs by directly minimizing the gradient boosting loss using stochastic gradient descent. As each weak learner is updated in parallel, this method diverges significantly from the gradient boosting framework.

An alternative line of research aims to arrive at a Bayesian soft GBM by making BART ``smoother''.
\citet{linero2017SBART} propose \textit{soft BART} (SBART), a model which replaces the standard decision trees in BART with soft trees. 
Empirically, fitting the resulting method is significantly slower than BART. 
\citet{maia2022GPBART} suggest putting Gaussian process priors on each leaf in a BART model, demonstrating that this method performs better than BART and SBART, but is also slower.

\section{METHODS}

\subsection{Variational soft trees}

\begin{figure}[t]
    \centering
    \resizebox{\linewidth}{!}{
    \begin{tabular}{cc}
        \subfloat[Soft tree]{\includegraphics[width=.4\linewidth]{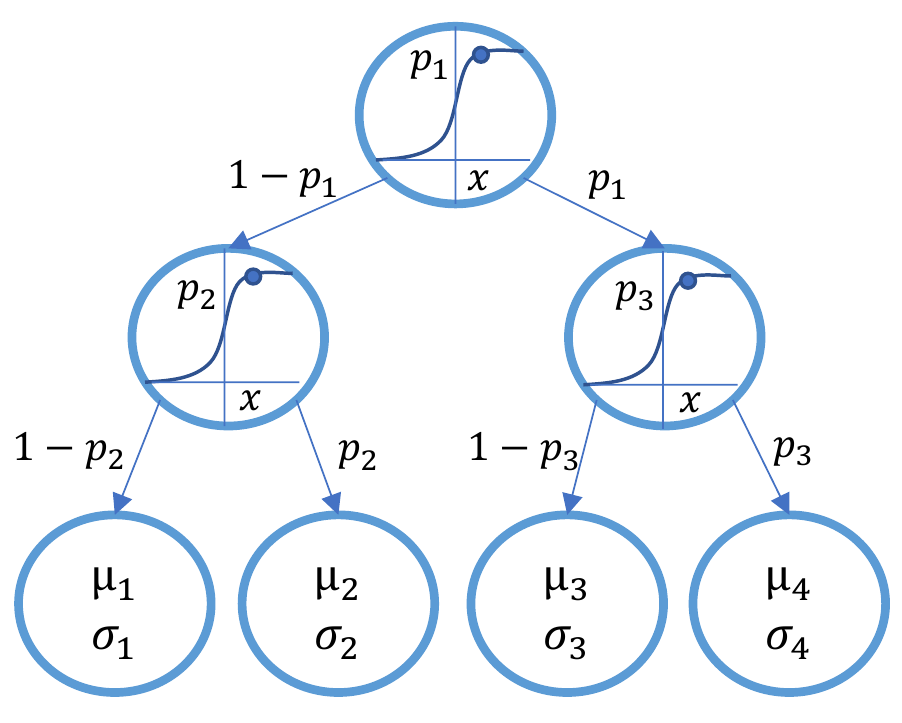}} &
        \subfloat[Bayesian soft tree]{\includegraphics[width=.4\linewidth]{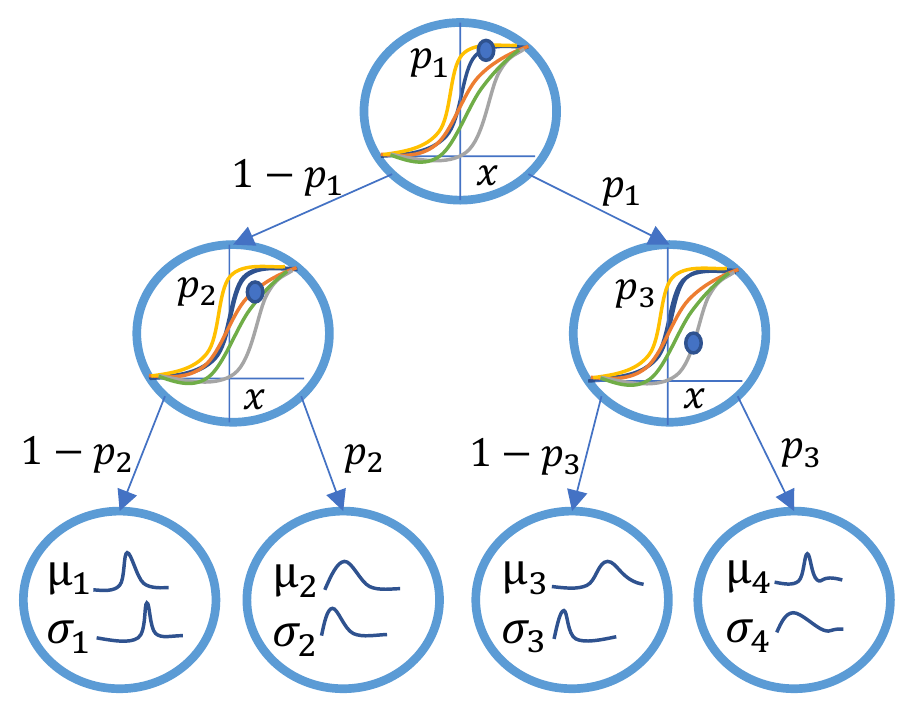}}
    \end{tabular}
    }
    \caption{Standard \vs Bayesian soft decision tree. In a standard soft tree each weight is assigned a fixed value, while in a Bayesian soft tree each weight is a distribution.}
    \label{fig:st_and_vst}
\end{figure}

In this section we propose a \emph{variational soft tree} (VST), a model that has the same architecture as a fixed depth soft decision tree (\cref{fig:st_and_vst}), but where we perform variational inference to approximate the full posterior distribution.
We then sample from the posterior approximation to compute the predictive distribution in \cref{eq:bayesian_prediction} by Monte Carlo integration.
We consider two variations of the variational soft tree: one with \emph{constant leaves} as in \citep{feng2020sgbm, frosst1017distilling, kontschieder2015deepneuralforest, Luo2021sdtr}, and another with \emph{linear leaves}, where each leaf outputs an affine function of the input. 
Linear leaves allow the model to represent piecewise linear functions, which are strictly more expressive than piecewise constant functions. 
In fact, a piecewise linear function can represent \emph{any} differentiable scalar function in the limit of infinite pieces by setting each piece to the derivative of the function.
Linear leaves use two additional parameter vectors per leaf, and in our experiments had a marginal effect on the computation cost.
To our knowledge, we are the first to consider soft decision trees with linear leaves.

\paragraph{Likelihood}
\label{sec:vist_likelihood}

We define the likelihood of the soft tree as 
\begin{align}
    \prob{y \g x, \theta} = \sum_{l \in L} \pr{l \g x, \phi} \prob{y \g x, l, \psi}, \label{eq:likelihood_1}
\end{align}
where $L$ is the set of tree leaves (terminal nodes) and $\theta = \set{\phi, \psi}$ are the parameters of the soft tree.

$\pr{l \g x, \phi}$, the first term inside the sum in \cref{eq:likelihood_1}, is the probability that the input $x$ is routed to the leaf $l$, defined as:
\begin{multline}
    \pr{l \g x, \phi} =
    \prod_{n \in \text{Path}\br{l}} 
    \Bigl[
    \pr{r \g n, x, \phi}^{n{\searrow}l} \\[-1.5em]
    \pr{\neg r \g n, x, \phi}^{l{\swarrow}n}
    \Bigr],
\end{multline}
where $\text{Path}\br{l}$ the set of tree nodes on the path from the root to the leaf $l$, $n{\searrow}l = 1$ if the leaf $l$ is in the \emph{right} subtree of node $n$, $l{\swarrow}n = 1$ if it is in the \emph{left} subtree, and $\pr{r \g n, x, \phi}$ is the probability that $x$ is routed to the right subtree at the node $n$, with $\pr{\neg r \g n, x, \phi} = 1 - \pr{r \g n, x, \phi}$.
We parametrize $\pr{r \g n, x, \phi}$ using a \emph{soft gating function} $f_\phi$:
\begin{align}
    \pr{r \g n, x, \phi} = f_\phi\br{x,n},
\end{align}
restricting $f_\phi\br{x,n} \in \sqbr{0, 1}$ for all $x$ and $n$.
In this work, we choose a simple linear gating function \citep{feng2020sgbm, frosst1017distilling, isroy2012softdecisiontree, Luo2021sdtr}:
\begin{align}
f_\phi\br{x, n} = \sigma\br{\beta \br{w_{n}^Tx+b_{n}}},
\end{align}
where $\sigma$ is the standard sigmoid function, $\beta$ is an inverse temperature parameter used to regularize the function during training \citep{frosst1017distilling}, and $\phi = \set{\br{w_n, b_n}}_{n \in N}$ are the parameters of the function for the set of tree nodes $N$.

The second term inside the sum in \cref{eq:likelihood_1}, $\prob{y \g x, l, \psi}$, is the predictive probability when the input $x$ is routed to the leaf $l$.
In this work, we consider two variations of this term: a \emph{constant leaf} model, and a \emph{linear leaf} model. 
In the constant leaf model, the output at each leaf is independent of the input: 
\begin{align}
\prob{y \g x, l, \psi} &=  \prob{y \g l, \psi} \\
&= \gaussianx{y}{\mu_l}{\softplus\br{\alpha_l}^2},
\end{align}
where $\psi = \set{(\mu_l, \alpha_l)}_{l \in L}$.
In the linear leaf model, on the other hand, the parameters of the Gaussian are linear in the input:
\begin{align}
\prob{y \g x, l, \psi} &= \gaussianx{y}{\mu_l\br{x}}{\sigma_l^2\br{x}} \\
\mu_l\br{x} &= w_l^T x + b_l \\
\sigma_l\br{x} &= \softplus\br{\hat{w}_l^T x + \hat{b}_l}
\end{align}
where $\psi = \set{(w_l, b_l, \hat{w}_l, \hat{b}_l)}_{l \in L}$ and $\softplus\br{x}=\text{log}(1+\text{exp}(x))$.
We use $\softplus$ functions to constrain the standard deviation parameters to remain strictly positive, while ensuring numerical stability \citep{blundell2016mfvibnn}.


\paragraph{Prior}

We use a spherical, zero centered Gaussian prior for the model parameters:
\begin{align}
    \prob{\theta} = \gaussianx{\theta}{0}{\tau^2 \eye}. 
\end{align}
Note that for a constant leaf model we put this prior on the inverse softplus of the standard deviation at each leaf, which restricts the variance to stay strictly positive.

\paragraph{Variational posterior}

We use the variational distribution proposed by \citet{tomczak2020lrvinn}, which is a Normal distribution with a low-rank covariance matrix:
\begin{align}
    q(\theta) = \gaussianx{\theta}{\mu}{\text{diag}[\sigma^2] + V V^T}
\end{align}
where 
$\sigma \in \mathbb{R}^p$ and $V$ is a matrix of rank $k$.
The hyperparameter $k$ determines the rank of the covariance matrix, and hence the expressivity of the posterior.
If $\theta \in \real^p$, choosing $k = p$ allows learning arbitrary covariance matrices, but increases the memory requirements and computational cost.
As a result, in our experiments we typically choose $k < p$. 
This choice of the variational posterior and the prior admits an analytical KL-divergence which is detailed in \cref{sec:vi} in the supplementary material.

\begin{algorithm}[tb]
\caption{Gradient boosted variational soft trees}
\label{alg:gbvst}
\KwIn{Prior parameters $a_\sigma$ and $b_\sigma$ \\ \hspace{3em} $T$ the number of soft trees in the GBM}
\KwData{$\set{X, y}$}
Fit variational soft tree $f_{\theta_1}$ to $\set{X, y}$ \\
\For{$t \in [2,\dots T]$}{
    Sample parameters for all models $f_{\theta_j}: \theta_{j} \sim q_j$ for $j = 1 \dots t-1$ \\
    Compute the residual vector: $r_t = y - \sum_{j=1}^{t-1} f_{\theta_j}\br{X}$ \\
    Fit the new model $f_{\theta_t}$ to $\set{X, r_t}$
}
Sample parameters $\theta = \set{\theta_j}_{j=1}^T$ where $\theta_{j} \sim q_j$ \\
Compute final residuals $r = y - F_\theta\br{X}$\\
Sample $\sigma^2 \sim \text{inverse-Gamma}(a_\sigma + n, b_\sigma + r^Tr)$ \\
\end{algorithm}

\subsection{Gradient boosted variational soft trees}

In order to increase the expressivity of the variational soft tree, we propose to use these models as weak learners in a gradient boosting machine that we refer to as a \emph{variational soft GBM} (VSGBM). More formally, the model is defined as
\begin{equation}
\label{eq:vsgbm}
    F_\theta\br{x} = \sum_{t=1}^T f_{\theta_t}\br{x} + \epsilon \quad \text{where} \quad \epsilon \sim \gaussianx{\epsilon}{0}{\sigma^2}
\end{equation}
where $f_{\theta_t}$ is a variational soft tree with parameters $\theta_t$, $T$ is the number of trees in the ensemble and $\sigma$ is a scale parameter. 
We consider variational soft trees $f_{\theta_t}$ that output the estimated mean value of the target and no standard deviation. 
Hence the variational soft GBM has a homoskedastic (input-independent) variance model.
Heteroskedastic (input-dependent) uncertainty could in principle be modeled by returning the predictive variances from the weak learners alongside the predictive means, and combining them upstream. In our early experiments this configuration proved to be costly to train (it doubles the number of model parameters), and made the training unstable (the gradient boosting loss had high variance). 
In prior work \citet{pratola2017hbart} train a separate ensemble to capture aleatoric uncertainty, likely to avoid this instability.

Inspired by \citet{he2020stochastictreeensemble}, we place an $\text{inverse-Gamma}(a_\sigma, b_\sigma)$ prior on $\sigma^2$, yielding an $\text{inverse-Gamma}(a_\sigma + n, b_\sigma + r^Tr)$ posterior, where $n$ is the number of data samples and $r = y - F_\theta\br{X}$ is the residual vector. 
The variational soft GBM is fit following the procedure described in \Cref{alg:gbvst}. 
Once trained, samples from the approximate posterior of the variational soft GBM are obtained by drawing parameters from the variational distribution of each variational soft tree model in the ensemble. 
The predictive distribution is then approximated by Monte Carlo integration of the likelihood as per \cref{eq:bayesian_prediction}, \ie by aggregating the predictions of the variational soft trees as defined by \cref{eq:vsgbm} for different posterior samples.

\section{EXPERIMENTS}

In this section we evaluate the variational soft tree and variational soft GBM on regression, out-of-distribution detection and multi-armed bandits. 
Although we are not measuring model calibration directly, like prior work we assume that model calibration and performance on these tasks are correlated.
When reporting results, we bold the highest score, as well as any score if its error bar and the highest score's error bar overlap, considering the difference to not be statistically significant.

\subsection{Regression}

\begin{table*}[tb]
\footnotesize
\scshape
\caption{Test log-likelihood of evaluated methods on regression datasets.
We find that variational soft decision trees (\textsc{VST}) and variational soft GBMs (\textsc{VSGBM}) perform strongly compared to baselines in terms of test log-likelihood, both methods out-performing them on 7/10 datasets.
}
\label{tab:vsdt_vsgbm_vs_baselines_ll}
\resizebox{\linewidth}{!}{
\begin{tabular}{@{}lcccccccccc@{}}
\toprule
Model   &   Power   & Protein & Boston  &   Naval   &   Yacht   &   Song    &   Concrete    & Energy &   Kin8nm  & Wine  \\ \midrule
Deep ensemble   &   -1.404 $\pm$ 0.001    & -1.422 $\pm$ 0.000 & -1.259 $\pm$ 0.007 &   -1.412 $\pm$ 0.000 &  -1.112 $\pm$ 0.009    & -1.415 $\pm$ 0.001  & -1.390 $\pm$ 0.001  & -1.677 $\pm$ 0.073 & -1.416 $\pm$ 0.000 & -1.725 $\pm$ 0.075 \\
\midrule
XBART (1 tree)  & -0.096 $\pm$ 0.003    &   -1.136 $\pm$ 0.003  & \textbf{-0.508 $\pm$ 0.042}   &  \textbf{-0.485 $\pm$ 0.419}    &  \textbf{\hphantom{-}1.110 $\pm$ 0.021}   &  -1.151 $\pm$ 0.006   & -0.793 $\pm$ 0.079 & -0.848 $\pm$ 0.567 & -1.153 $\pm$ 0.014    & -1.250 $\pm$ 0.008 \\
SGLB (1 tree)  &   -1.425 $\pm$ 0.027  & -1.499 $\pm$ 0.002    & -1.361 $\pm$ 0.044 & -1.410 $\pm$ 0.009 & -0.891 $\pm$ 0.047    &  -1.305 $\pm$ 0.007   & -1.748 $\pm$ 0.050   & -1.078 $\pm$ 0.006  & -1.659 $\pm$ 0.015    & -1.559 $\pm$ 0.033 \\
VST (const.)   &   -0.032 $\pm$ 0.010  & -0.719 $\pm$ 0.005    &   -0.920 $\pm$ 0.002  &   -0.555 $\pm$ 0.088  & \hphantom{-}0.193 $\pm$ 0.024 & -0.733 $\pm$ 0.010    & -0.999 $\pm$ 0.006    & -0.710 $\pm$ 0.280  & -0.877 $\pm$ 0.014    & \textbf{\hphantom{-}1.458 $\pm$ 0.615} \\
VST (linear)  & \textbf{\hphantom{-}0.024 $\pm$ 0.002}    &   \textbf{-0.631 $\pm$ 0.011} & -0.682 $\pm$ 0.004 & \textbf{-0.335 $\pm$ 0.026} & -0.754 $\pm$ 0.005   &  \textbf{-0.579 $\pm$ 0.002}  & \textbf{-0.461 $\pm$ 0.070}   &  \textbf{\hphantom{-}0.377 $\pm$ 0.123}   &  \textbf{-0.707 $\pm$ 0.012}    & \textbf{\hphantom{-}3.067 $\pm$ 1.168} \\
\midrule
XBART   & \textbf{-0.001 $\pm$ 0.006}   & -1.136 $\pm$ 0.003    & \textbf{-0.462 $\pm$ 0.054}  & \hphantom{-}0.008 $\pm$ 0.130    & \textbf{\hphantom{-}1.116 $\pm$ 0.023}    & -1.018 $\pm$ 0.002    & -0.581 $\pm$ 0.024  & -0.462 $\pm$ 0.054    & -1.057 $\pm$ 0.007    & -1.251 $\pm$ 0.010 \\
SGLB    &  -0.500 $\pm$ 0.001   & -1.204 $\pm$ 0.002    & \textbf{-0.504 $\pm$ 0.013}   & -0.877 $\pm$ 0.004  & -0.018 $\pm$ 0.009    & -0.907 $\pm$ 0.001  & -0.931 $\pm$ 0.010    & -0.232 $\pm$ 0.009 & -1.125 $\pm$ 0.002    & -1.220 $\pm$ 0.006 \\ 
VSGBM (const.) & \textbf{-0.002 $\pm$ 0.011}   & \textbf{-1.039 $\pm$ 0.005}   & \textbf{-0.567 $\pm$ 0.062} & \textbf{\hphantom{-}0.573 $\pm$ 0.311}    & \hphantom{-}0.045 $\pm$ 0.912    & -0.933 $\pm$ 0.008 & \textbf{-0.465 $\pm$ 0.040}   & -0.227 $\pm$ 0.038    & -0.457 $\pm$ 0.021    & -1.192 $\pm$ 0.007 \\
VSGBM (linear)    & \textbf{-0.015 $\pm$ 0.011} & \textbf{-1.051 $\pm$ 0.012}  & \textbf{-0.462 $\pm$ 0.054}    & \textbf{\hphantom{-}0.542 $\pm$ 0.295}    & -0.033 $\pm$ 0.049    & \textbf{-0.729 $\pm$ 0.019}    & -0.591 $\pm$ 0.079    & \textbf{\hphantom{-}0.057 $\pm$ 0.134}    & \textbf{-0.363 $\pm$ 0.031}    & \textbf{-1.167 $\pm$ 0.004} \\
\bottomrule
\end{tabular}
}
\end{table*}

\begin{figure*}[tb]
    \centering
    \resizebox{.9\linewidth}{!}{
    \begin{tabular}{ccc}
    \subfloat[Variational soft tree]{\includegraphics[width=0.33\linewidth]{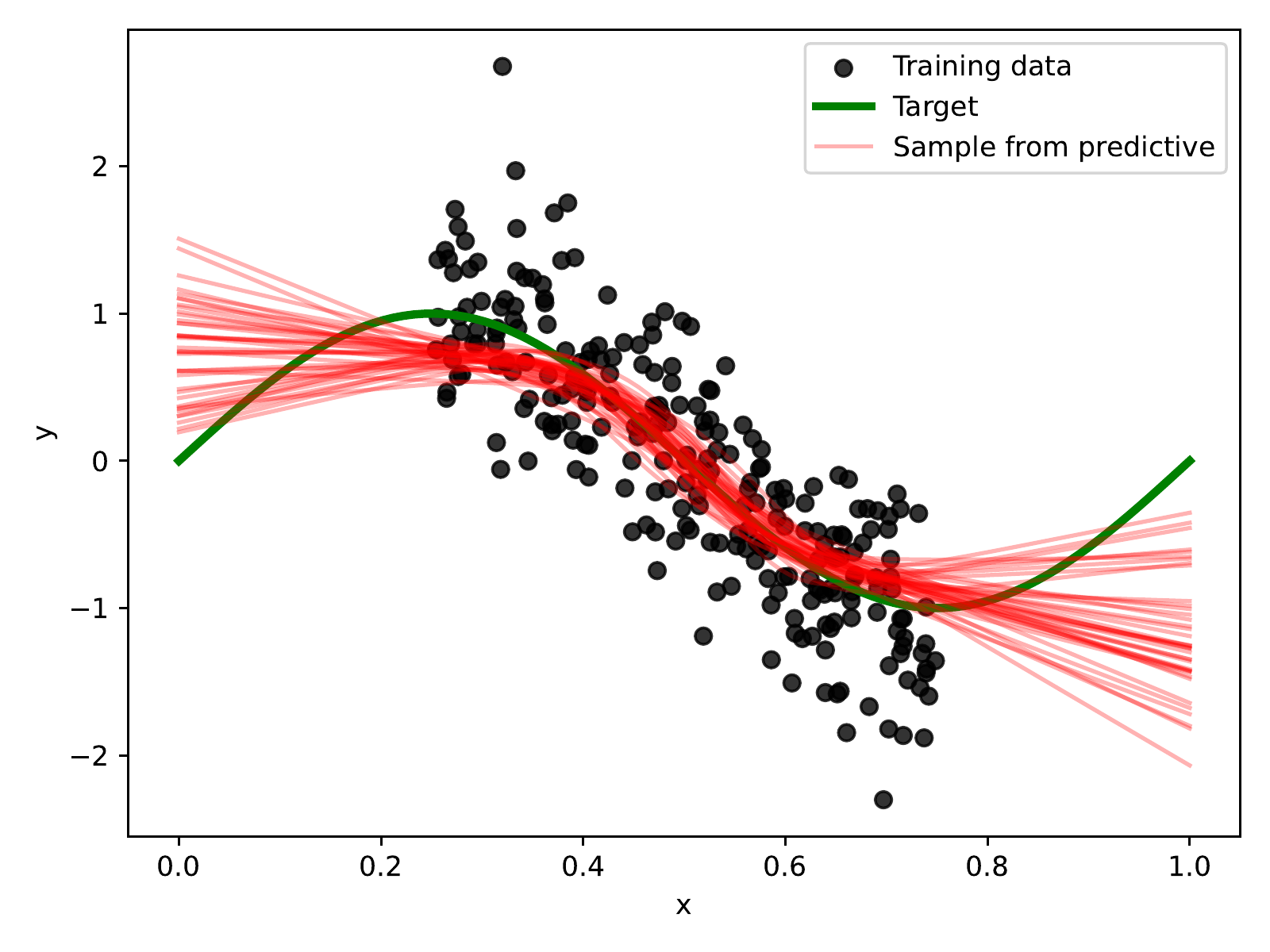}} &
    \subfloat[XBART]{\includegraphics[width=0.33\linewidth]{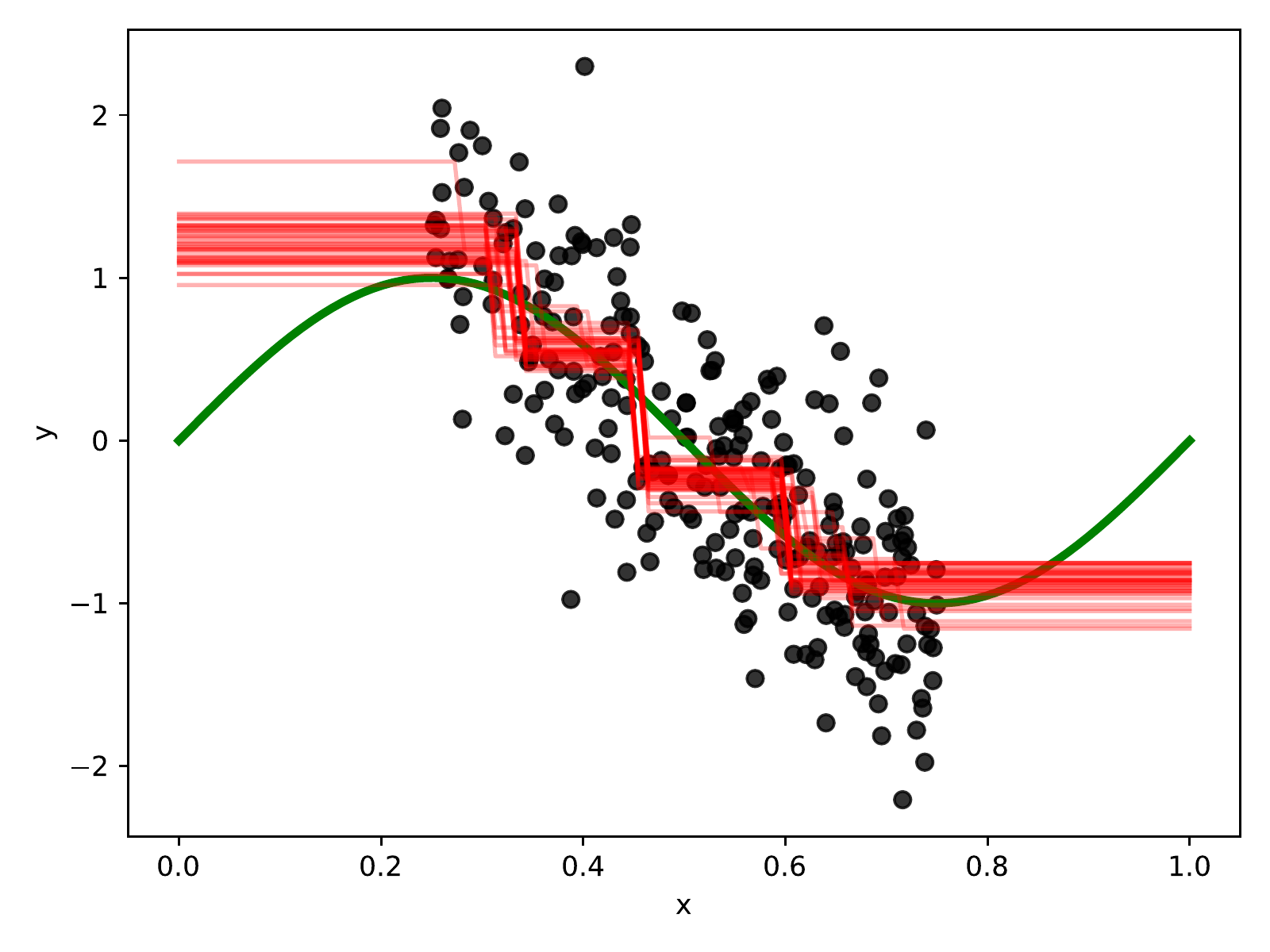}} &
    \subfloat[SGLB]{\includegraphics[width=0.33\linewidth]{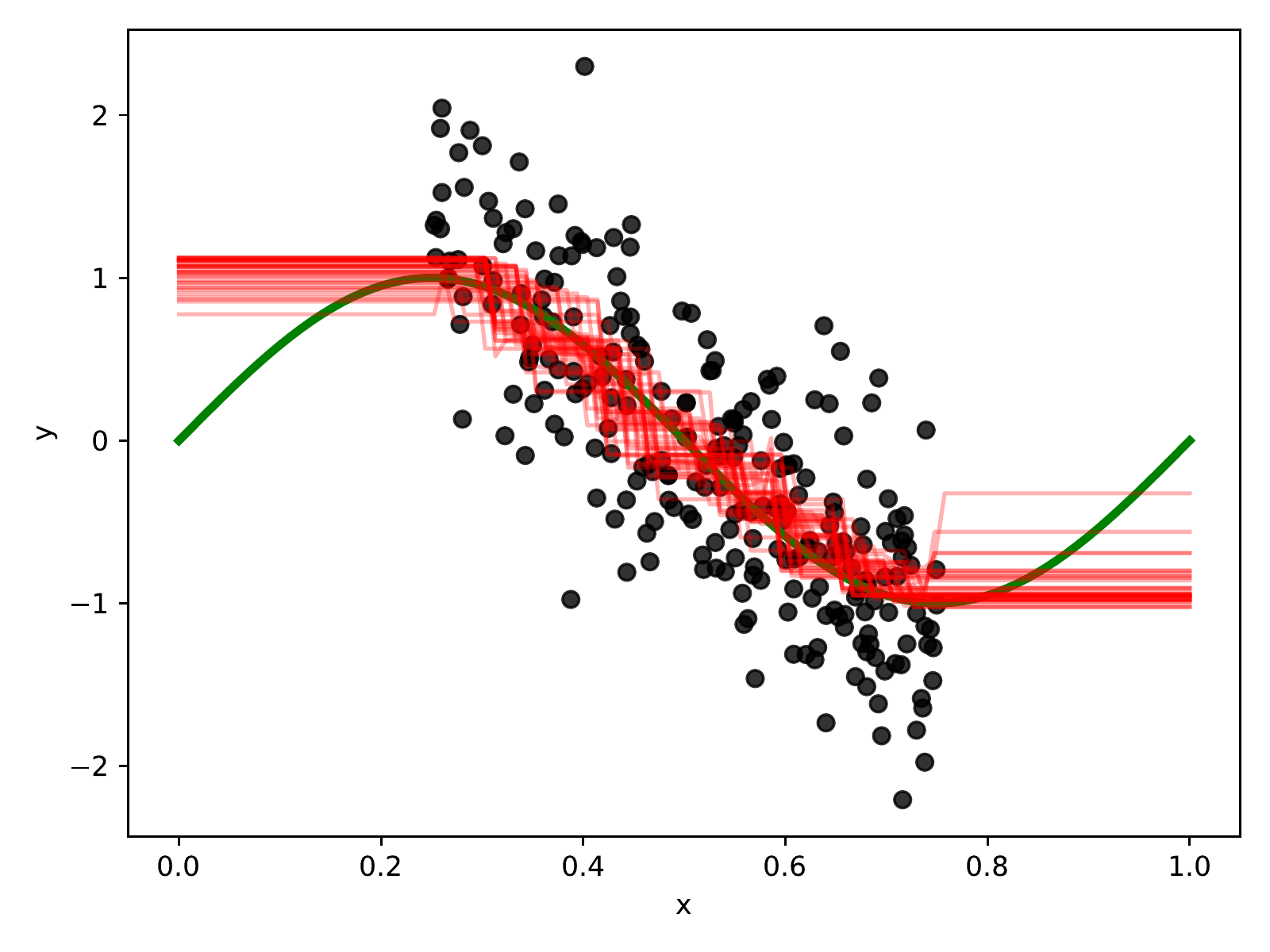}}
    \end{tabular}
    }
    \caption{Variational soft trees (VST) show high uncertainty in the tails outside of the data support while closely agreeing with the mean inside the support. Sampled function show more diversity in tails with VSTs than with XBART and SGLB. The VST uses linear leaves, while XBART and SGLB use constant leaves.}
    \label{fig:tail_uncertainty}
\end{figure*}

We consider a suite of regression datasets used by \citet{malinin2021uncertGBM} described in \cref{tab:dataset_description} in supplementary material. 
We perform 10-fold cross validation on the training data and report the mean and standard deviation of the test log-likelihood and root mean square error (RMSE) of our models across the folds.
We compare our variational soft decision trees (\textsc{VST}) and variational soft GBM (\textsc{VSGBM}) with two Bayesian GBMs (XBART \citep{he2020stochastictreeensemble} and SGLB \citep{ustimenko2020sglb}), an ensemble of neural networks, and XGBoost \citep{Chen2016XGBOOST}. 
When comparing models to VSTs, we consider baselines with an ensemble size of one, and otherwise multiple ensemble members when comparing with the VSGBM. 
The only exception is the ensemble of deep neural networks for which we always use an ensemble of ten models to measure epistemic uncertainty.

\paragraph{Variational soft decision tree}

We find that our model performs well in terms of test log-likelihood and RMSE across the considered datasets (see \cref{tab:vsdt_vsgbm_vs_baselines_ll,tab:vsdt_vsgbm_vs_baselines_rmse}), which confirms that soft decision trees are well-suited to tabular data \citep{Luo2021sdtr}.
VSTs outperform the baselines on 7/10 datasets in terms of log-likelihood, and 3/10 datasets in terms of RMSE, also matching the log-likehood and RMSE of best performing baselines on 1/10 and 3/10 datasets respectively. 
In addition, soft decision trees with linear leaves perform better than equivalent constant leaf models on 8/10 datasets in terms of log-likelihood and 5/10 in terms of RMSE, bringing evidence that linear leaf models are more expressive than constant leaf models.

VSTs are less competitive when trained on small datasets like \textit{boston} (506 training examples) and \textit{yacht} (308 training examples).
We hypothesize that in these cases the model underfits the data due to the choice of the prior.
The parameters of the prior were chosen by hyper-parameter tuning with the validation ELBO objective, and the underfitting behavior could be an artifact of this tuning procedure.
An alternative hypothesis is that a normal prior is simply ill-suited to this model. 
Choosing priors for complex, deep models is difficult, and is an active research direction in Bayesian deep learning 
\citep{FlamShepherd2017MappingGP, fortuin2021bnnpriorrevisited, Krishnan_Subedar_Tickoo_2020, Vladimirova2018understandingpriorsbnn}.

\paragraph{Variational soft GBM}

VSGBMs perform competitively in terms of log-likelihood compared to the baselines (see \cref{tab:vsdt_vsgbm_vs_baselines_ll,tab:vsdt_vsgbm_vs_baselines_rmse}), yielding the best log-likelihood on 7/10 datasets. 
The VSGBMs perform on par with baselines on \emph{power} and \emph{boston}, worse on \emph{yacht} but better on the other datasets.
However, VSGBMs are less competitive in terms of RMSE when compared to XGBoost that yields smaller RMSE on all but three datasets.
Compared to other baselines, VSGBM consistently outperforms the deep ensemble, and performs favorably when compared to SGLB and XBART.
In addition, we find that VSGBMs with linear leaves perform slightly better than equivalent constant leaf models, outperforming them on 4/10 datasets in terms of log-likelihood and 3/10 in terms of RMSE, while also matching the log-likelihod and RMSE of constant leaves on 5/10 and 6/10 datasets respectively.


Compared to a VST, a VSGBM performs better in terms of RMSE on all datasets but two where both models score equally well. 
However, VSTs obtain higher log-likelihoods than VSGBMs on 5/10 datasets, equivalent scores on 2/10 and performs worse on 3/10.
We explain this by the simpler variance model used in the VSGBM, which does not allow for heteroskedastic (input dependent) uncertainty.

\paragraph{Uncertainty estimation visualization}

An important property of a reliable predictive model is its increased predictive uncertainty for inputs in regions with little training data.
To understand if VSTs demonstrate this behavior, we plot functions sampled from the variational posterior for simple synthetic datasets, and compare them to functions sampled from XBART and SGLB posteriors.
We plot the sampled functions in \cref{fig:tail_uncertainty,fig:app_inbetween_uncertainty}.

We find that VSTs indeed capture the uncertainty outside of the support of the data better than both XBART and SGLB: functions agree with the true mean within the support of the data, but diverge outside of it.

\subsection{OOD detection}

\begin{table*}[tb]
\footnotesize
\scshape
\caption{OOD detection AUROC scores with variational soft trees (\textsc{VST}) and variational soft GBMs (\textsc{VSGBM}) against baselines (higher is better). 
Variational soft decision trees and variational soft GBMs perform well on out of distribution detection, yielding the best score on 6/10 and 5/10 datasets respectively.
}
\label{tab:OODDetection}
\resizebox{\linewidth}{!}{
\begin{tabular}{@{}lcccccccccc@{}}
\toprule
Model   &   Power   & Protein & Boston  &   Naval   &   Yacht   &   Song    &   Concrete    & Energy &   Kin8nm  & Wine  \\ \midrule
Deep ensemble   & 0.540 $\pm$ 0.015 & 0.504 $\pm$ 0.003 & 0.545 $\pm$ 0.023 & \textbf{1.000 $\pm$ 0.000}    & \textbf{0.691 $\pm$ 0.088}    & 0.510 $\pm$ 0.007  & \textbf{0.767 $\pm$ 0.042} & 0.554 $\pm$ 0.036   & 0.508 $\pm$ 0.005 & 0.547 $\pm$ 0.019 \\
\midrule
XBART (1 tree)  & 0.562 $\pm$ 0.018 & 0.684 $\pm$ 0.015 & \textbf{0.656 $\pm$ 0.042} & 0.578 $\pm$ 0.050 & 0.603 $\pm$ 0.021 & \textbf{0.608 $\pm$ 0.033}    & 0.671 $\pm$ 0.028 & 0.852 $\pm$ 0.141   & 0.525 $\pm$ 0.010 & 0.548 $\pm$ 0.016 \\
SGLB (1 tree)   &  \textbf{0.593 $\pm$ 0.025}  &  0.760 $\pm$ 0.018    & \textbf{0.732 $\pm$ 0.039}  & 0.770 $\pm$ 0.021 & 0.607 $\pm$ 0.035 & \textbf{0.589 $\pm$ 0.035} & 0.723 $\pm$ 0.018 & 0.969 $\pm$ 0.003   & 0.519 $\pm$ 0.005 & 0.571 $\pm$ 0.021 \\
VST (const.)  &  \textbf{0.591 $\pm$ 0.025} & 0.866 $\pm$ 0.018    & \textbf{0.712 $\pm$ 0.020} & 0.884 $\pm$ 0.024   & \textbf{0.718 $\pm$ 0.048}    & \textbf{0.560 $\pm$ 0.025} & 0.677 $\pm$ 0.049   & 0.863 $\pm$ 0.045 & 0.529 $\pm$ 0.016 & \textbf{0.796 $\pm$ 0.010} \\
VST (linear)  & \textbf{0.618 $\pm$ 0.024}    & \textbf{0.920 $\pm$ 0.019}    & \textbf{0.698 $\pm$ 0.020}   & \textbf{1.000 $\pm$ 0.000}  & 0.655 $\pm$ 0.015 & \textbf{0.601 $\pm$ 0.015}    & \textbf{0.840 $\pm$ 0.013} & \textbf{0.996 $\pm$ 0.003}    & \textbf{0.586 $\pm$ 0.011}    & 0.742 $\pm$ 0.036 \\
\midrule
XBART   & 0.572 $\pm$ 0.011 & 0.678 $\pm$ 0.016 & \textbf{0.741 $\pm$ 0.041} &  0.912 $\pm$ 0.033 & \textbf{0.607 $\pm$ 0.026} &  0.565 $\pm$ 0.013 & \textbf{0.818 $\pm$ 0.020} & 0.837 $\pm$ 0.146 & 0.541 $\pm$ 0.011 & 0.549 $\pm$ 0.021 \\
SGLB    & \textbf{0.600 $\pm$ 0.011}    & 0.874 $\pm$ 0.006  & 0.738 $\pm$ 0.027 & 0.907 $\pm$ 0.011 & \textbf{0.662 $\pm$ 0.036} & \textbf{0.697 $\pm$ 0.029} &  0.763 $\pm$ 0.024 & 0.973 $\pm$ 0.007 & 0.508 $\pm$ 0.004 &  0.613 $\pm$ 0.019  \\
VSGBM (const.) & 0.569 $\pm$ 0.018 & 0.890 $\pm$ 0.011 & \textbf{0.775 $\pm$ 0.101}    & \textbf{1.000 $\pm$ 0.000}    & \textbf{0.663 $\pm$ 0.055}    & 0.586 $\pm$ 0.014    & 0.666 $\pm$ 0.033    & 0.901 $\pm$ 0.018    & 0.562 $\pm$ 0.015    & 0.625 $\pm$ 0.019   \\
VSGBM (linear)    & 0.571 $\pm$ 0.010 & \textbf{0.950 $\pm$ 0.016}    & \textbf{0.795 $\pm$ 0.025}     & \textbf{1.000 $\pm$ 0.000}     &  \textbf{0.618 $\pm$ 0.020}     & 0.602 $\pm$ 0.006     & 0.763 $\pm$ 0.023     & \textbf{0.993 $\pm$ 0.008}     & \textbf{0.608 $\pm$ 0.012}     & \textbf{0.688 $\pm$ 0.027} \\
\bottomrule
\end{tabular}
}
\end{table*}

\begin{figure}[tb]
    \centering
    \includegraphics[width=.85\linewidth]{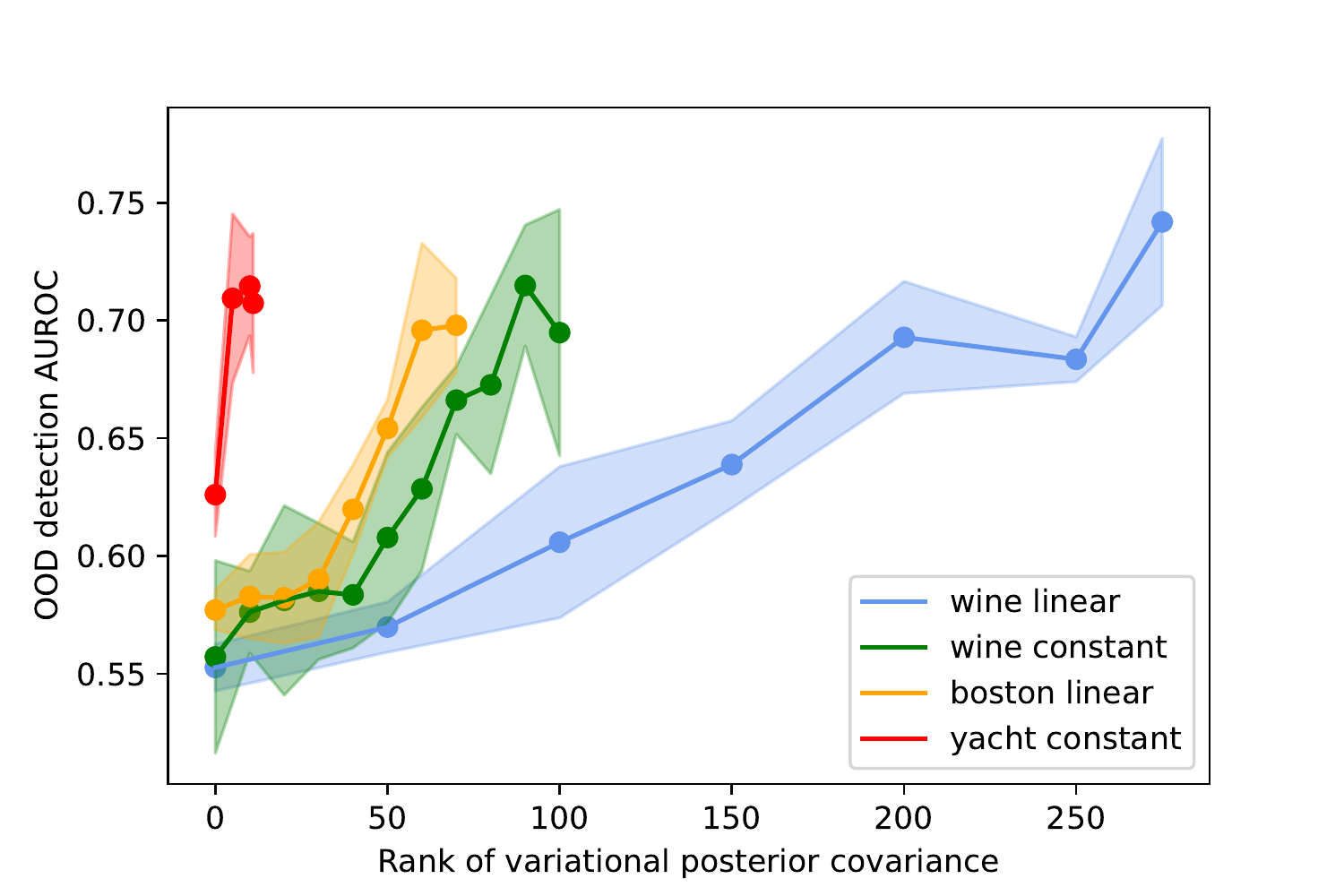}
    \caption{OOD detection AUROC score vs.\ the rank of the variational posterior covariance matrix in variational soft trees (\textsc{VST}). The quality of the OOD detection generally increases with the rank of the approximate posterior covariance. As we consider a large number of different model and dataset configurations, we only present the ones for which the increase in AUROC is most significant. The rightmost point of each plot corresponds to a full rank covariance approximation.
    }
    \label{fig:ood_detect_vs_cov_rk}
\end{figure}

Out-of-distribution (OOD) data is data that is sampled from a distribution that is sufficiently different from the training data distribution.
It has been observed in prior work \citep{ma2021FVISPG, tomczak2020lrvinn} that an effective Bayesian model will show high epistemic uncertainty when presented with OOD data, so we further evaluate our models by testing if their epistemic uncertainty is predictive of OOD data. 

Following the setup of \citet{malinin2021uncertGBM}, we define OOD data by selecting a subset of data from a different dataset.
To perform OOD detection, we compute the epistemic uncertainty of each sample under our models and evaluate how well a single splitting threshold separates the epistemic uncertainty of in-distribution and OOD samples, as proposed by \citet{Mukhoti2021ddu}.
Epistemic uncertainty is measured as the variance of the mean prediction with respect to samples from the posterior or the ensemble.
The detection performance is quantified using the area under the receiver operating characteristic curve (AUROC).
We report the mean and standard deviation of the AUROC of each model across 10 folds of cross-validation. 
As before, we compare our VSTs and VSGBMs with XBART, SGLB and a deep ensemble.

\begin{figure*}[tb]
    \centering
    \resizebox{\linewidth}{!}{
    \begin{tabular}{cccc}
    \subfloat[Financial]{\includegraphics[width=0.25\linewidth]{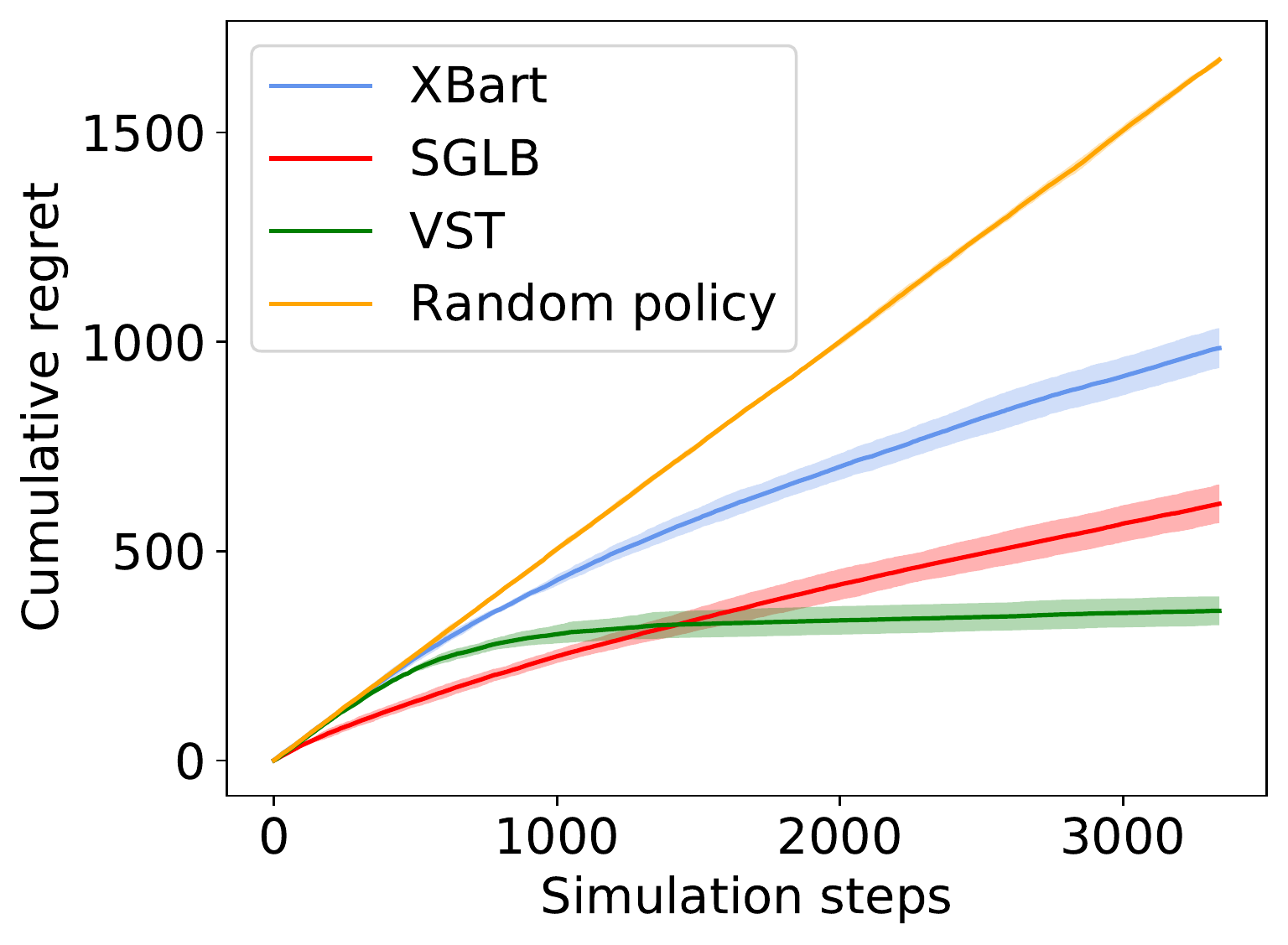}} &
    \subfloat[Mushroom]{\includegraphics[width=0.25\linewidth]{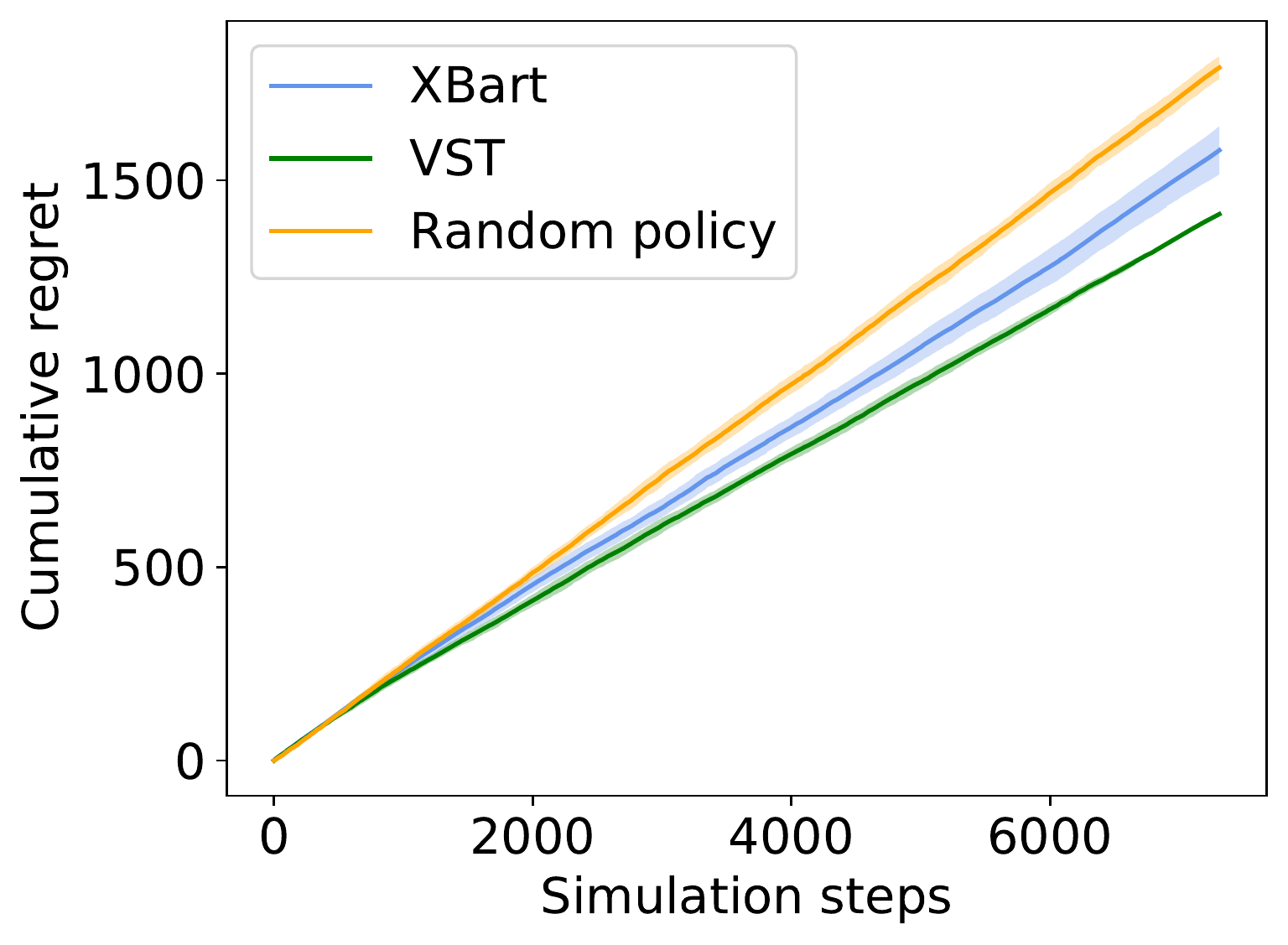}} &
    \subfloat[Jester]{\includegraphics[width=0.25\linewidth]{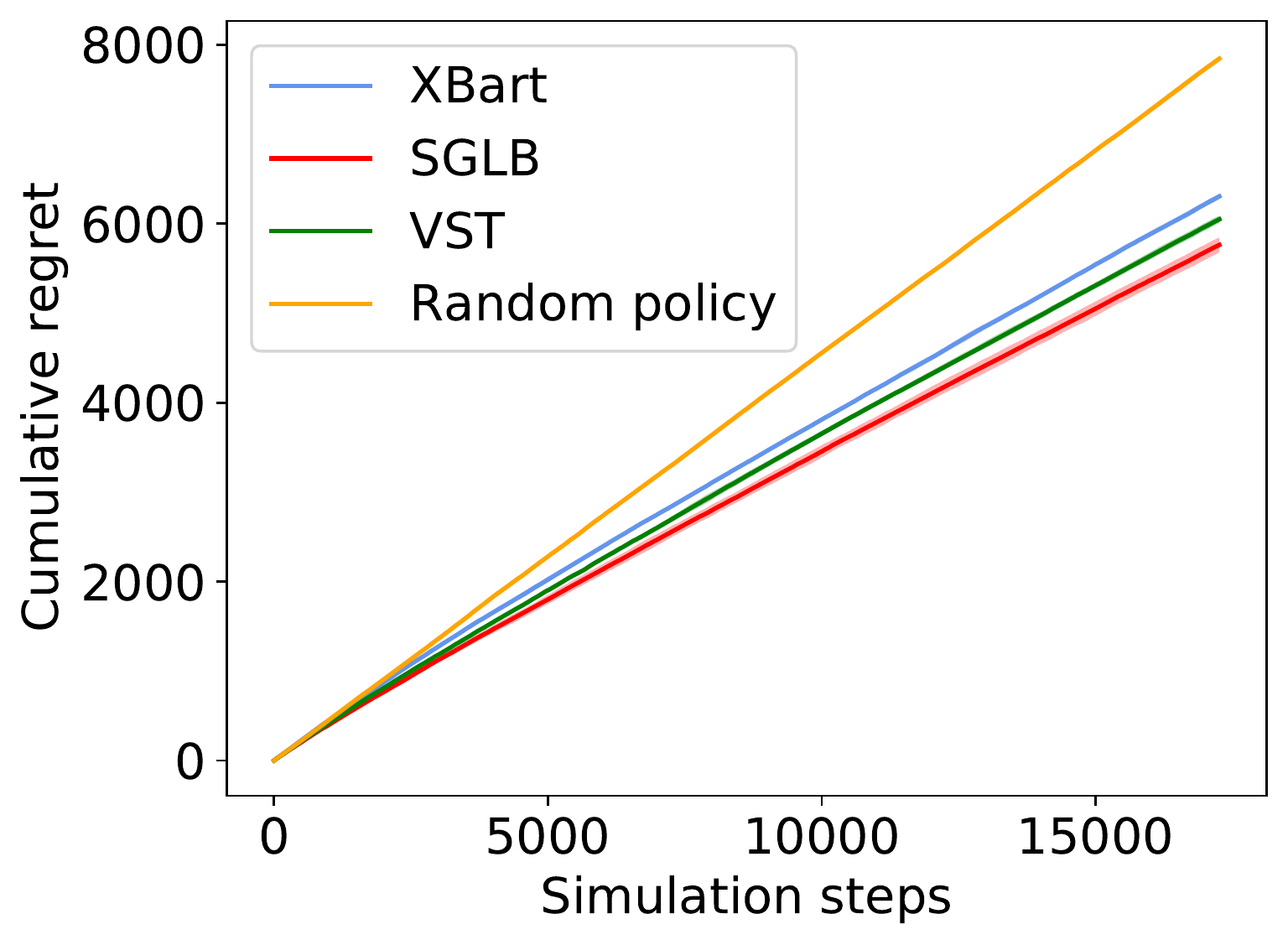}} &
    \subfloat[Exploration]{\includegraphics[width=0.25\linewidth]{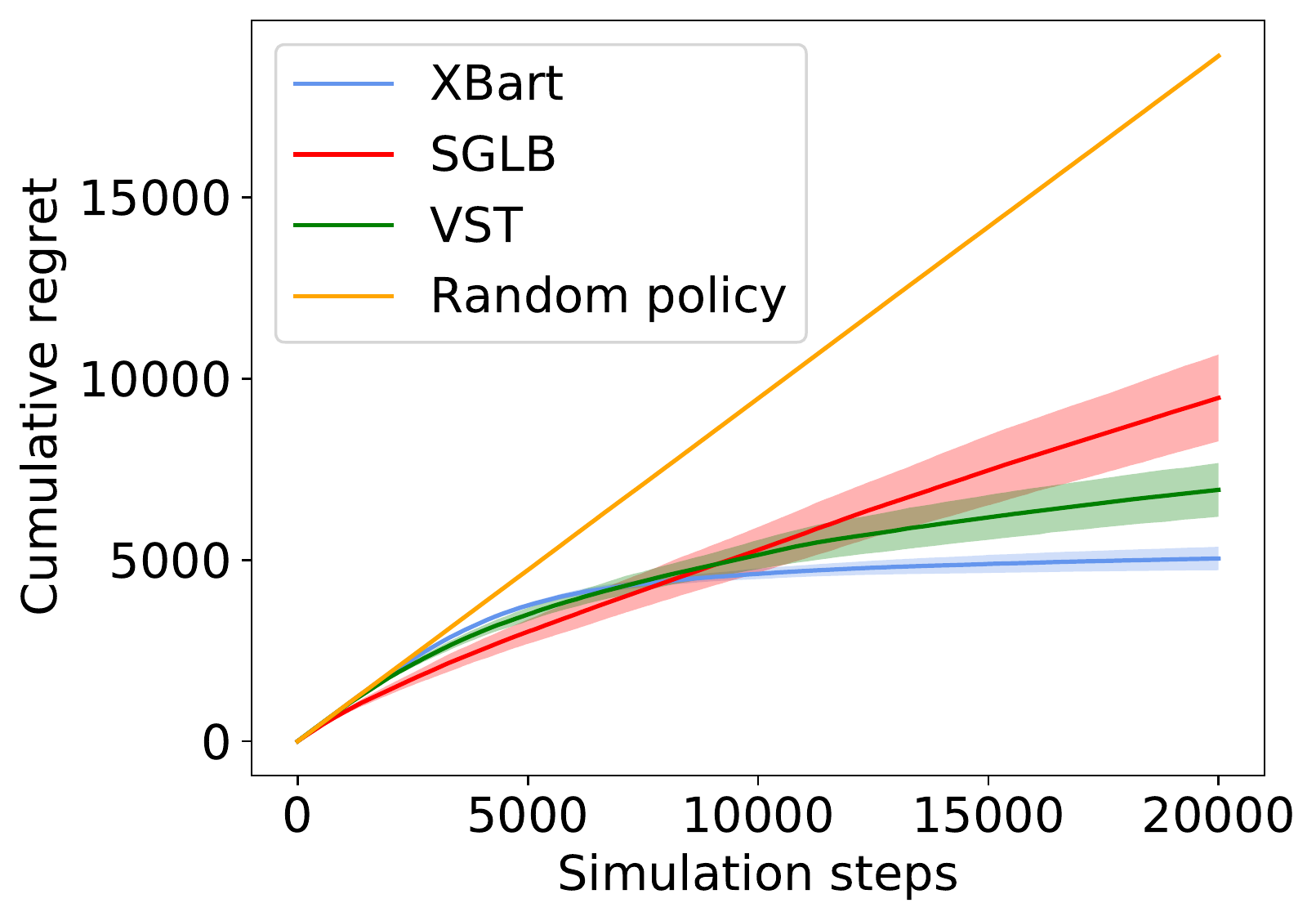}}
    \end{tabular}
    }
    \caption{Bandit results. We find that the variational soft tree (\textsc{VST}) converges to near zero instantaneous regret on the financial bandit and performs competitively compared to baselines in the bandit setting.}
    \label{fig:bandit_results}
\end{figure*}

We find that variational soft decision trees perform well on OOD detection, outperforming baselines on 6/10 datasets and tying on others (see \cref{tab:OODDetection}). 
VSGBMs also perform well, yielding the best scores on 5/10 datasets and additionally matching best baselines on 2/10 (see \cref{tab:OODDetection}). 
However, the improvements over the top baselines are less significant than with variational soft decision trees.
As expected, we find that increasing the rank of the approximate variational posterior increases the OOD performance, although the significance of the improvement depends on the dataset (see \cref{fig:ood_detect_vs_cov_rk}). 
This effect is most prominent with variational soft decision trees, and is less so in VSGBMs. 

\subsection{Bandits}

\begin{table}
\footnotesize
\scshape
\caption{Cumulative regret of the bandits using variational soft trees (\textsc{VST}) \vs baselines. We find that VST performs competitively compared to baselines, achieving the lowest cumulative regret on 2/4 bandits.}
\label{tab:bandit_results}
\resizebox{\linewidth}{!}{
\begin{tabular}{@{}lcccc@{}}
\toprule
Bandit      & financial     & exploration   & mushroom      & jester    \\ \midrule
Random policy   & 1682.05 $\pm$ 20.23 & 18904.95 $\pm$ 24.37\hphantom{00}  & 1791.28 $\pm$ 29.61 & 7845.28 $\pm$ 26.99\\
VST &   \textbf{\hphantom{0}357.47 $\pm$ 34.32} & \hphantom{0}7054.60 $\pm$ 748.79\hphantom{0}  & \textbf{1412.55 $\pm$ 5.64\hphantom{0}} & 6048.99 $\pm$ 42.26\\
SGLB    &   \hphantom{0}613.12 $\pm$ 45.96 &  \hphantom{0}9469.07 $\pm$ 1197.60   & doesn't work    & \textbf{5762.89 $\pm$ 82.10}\\
XBART   &   \hphantom{0}984.86 $\pm$ 47.59 & \textbf{5039.93 $\pm$ 322.63} & 1577.70 $\pm$ 61.56  & 6305.43 $\pm$ 31.97\\ \bottomrule
\end{tabular}
}
\end{table}

We further evaluate the VST by using it in Thompson sampling \citep{thompson1933TS} on four different multi-armed bandit problems (see \cref{app:bandit_datasets} in the supplementary material for more detail on the datasets), comparing to XBART and SGLB as before.
We do not experiment with VSGBMs in the bandit setting due to computational cost. 
In this setting, the model has to be re-trained after every update, and training a VSGBM is expensive: boosting is sequential by nature, and VST weak learners require a number of gradient descent iterations to converge.
Each bandit simulation is run five times with different random seeds, and we report the mean and the standard deviation of the cumulative regret across the simulations. 

The predictive uncertainty of VST is competitive in this setting as well, as shown in \cref{tab:bandit_results}.
The VST obtains the lowest cumulative regret on the \textit{mushroom} and \textit{financial} datasets, converging to near zero instantaneous regret on the latter (see \cref{fig:bandit_results}). 
 On the \textit{jester} dataset, the VST demonstrates second best cumulative regret, although the performance of all three models is comparable to the random policy.
We explain this by the difficulty of the regression problem in this particular dataset. 
Finally, just like on \textit{jester}, the VST shows second best cumulative regret on the \textit{exploration} dataset.
\section{DISCUSSION}

In this work we proposed to perform variational inference to approximate the Bayesian posterior of a soft decision tree.
We use a variational distribution that allows trading off memory and computation for a richer posterior approximation.
Using soft decision trees models as weak learners, we build a variational soft GBM that is more expressive than the individual trees, but preserves their well-calibrated predictive uncertainty.
We find that both variational soft trees and variational soft GBMs perform well on tabular data compared to strong baselines, and provide useful uncertainty estimates for OOD detection and multi-armed bandits.

An important direction for future work is to speed up the training of variational soft trees and GBMs.
This is especially important when working in the low-data regime (${<}10^4$ samples), where our models are slower than the baseline based on standard decision trees, and in the context of multi-armed bandits, where the model needs to be updated repeatedly.
For the latter, an alternative direction is to adapt our models for continual learning by implementing an efficient Bayesian update, which would accelerate sequential learning tasks like multi-armed bandits, Bayesian optimization and active learning.
Finally, addapting VST and VSGBM to the classification setting is an interesting direction for future work.
Classification with tabular data is important in many applications, many of which might also benefit from well-calibrated predictive uncertainties \citep{Ebbehoj2021TLforHCML, computers10020024}.


\bibliography{main.bib}

\begin{thebibliography}{}

\bibitem[Abdullah et~al., 2022]{medicineBDL}
Abdullah, A., Hassan, M., and Mustafa, Y. (2022).
\newblock A review on {B}ayesian deep learning in healthcare: Applications and
  challenges.
\newblock {\em IEEE Access}, 10:1--1.

\bibitem[Ahmeto{\u{g}}lu et~al., 2018]{Ahmetoglu2018csdt}
Ahmeto{\u{g}}lu, A., {\.{I}}rsoy, O., and Alpayd{\i}n, E. (2018).
\newblock Convolutional soft decision trees.
\newblock In K{\r{u}}rkov{\'a}, V., Manolopoulos, Y., Hammer, B., Iliadis, L.,
  and Maglogiannis, I., editors, {\em Artificial Neural Networks and Machine
  Learning -- ICANN 2018}, pages 134--141, Cham. Springer International
  Publishing.

\bibitem[Bew et~al., 2019]{analystRecommendationBML}
Bew, D., Harvey, C.~R., Ledford, A., Radnor, S., and Sinclair, A. (2019).
\newblock Modeling analysts{\textquoteright} recommendations via {B}ayesian
  machine learning.
\newblock {\em The Journal of Financial Data Science}, 1(1):75--98.

\bibitem[Bishop and Svensen, 2012]{bishop2002BayesianHME}
Bishop, C.~M. and Svensen, M. (2012).
\newblock {B}ayesian hierarchical mixtures of experts.

\bibitem[Blundell et~al., 2015]{blundell2016mfvibnn}
Blundell, C., Cornebise, J., Kavukcuoglu, K., and Wierstra, D. (2015).
\newblock Weight uncertainty in neural networks.

\bibitem[Bojer and Meldgaard, 2021]{Bojer_2021}
Bojer, C.~S. and Meldgaard, J.~P. (2021).
\newblock Kaggle forecasting competitions: An overlooked learning opportunity.
\newblock {\em International Journal of Forecasting}, 37(2):587--603.

\bibitem[Borisov et~al., 2021]{borisov2021DLTDsurvey}
Borisov, V., Leemann, T., Seßler, K., Haug, J., Pawelczyk, M., and Kasneci, G.
  (2021).
\newblock Deep neural networks and tabular data: A survey.

\bibitem[Bradbury et~al., 2018]{jax2018github}
Bradbury, J., Frostig, R., Hawkins, P., Johnson, M.~J., Leary, C., Maclaurin,
  D., Necula, G., Paszke, A., Vander{P}las, J., Wanderman-{M}ilne, S., and
  Zhang, Q. (2018).
\newblock {JAX}: Composable transformations of {P}ython+{N}um{P}y programs.

\bibitem[Chen and Guestrin, 2016]{Chen2016XGBOOST}
Chen, T. and Guestrin, C. (2016).
\newblock {XGBoost}: {A} scalable tree boosting system.
\newblock In {\em Proceedings of the 22nd {ACM} {SIGKDD} International
  Conference on Knowledge Discovery and Data Mining, San Francisco, CA, USA,
  August 13-17, 2016}, pages 785--794.

\bibitem[Chipman et~al., 1998]{chipman1998BayesianCART}
Chipman, H.~A., George, E.~I., and McCulloch, R.~E. (1998).
\newblock {B}ayesian cart model search.
\newblock {\em Journal of the American Statistical Association},
  93(443):935--948.

\bibitem[Clements et~al., 2020]{clements2020DLcreditmonitoring}
Clements, J.~M., Xu, D., Yousefi, N., and Efimov, D. (2020).
\newblock Sequential deep learning for credit risk monitoring with tabular
  financial data.

\bibitem[Cobb and Jalaian, 2020]{cobb2020hmc}
Cobb, A.~D. and Jalaian, B. (2020).
\newblock Scaling {H}amiltonian {M}onte {C}arlo inference for {B}ayesian neural
  networks with symmetric splitting.

\bibitem[Coppens et~al., 2019]{Coppens2019DistillingDR}
Coppens, Y., Efthymiadis, K., Lenaerts, T., and Now{\'e}, A. (2019).
\newblock Distilling deep reinforcement learning policies in soft decision
  trees.
\newblock In {\em IJCAI 2019}.

\bibitem[Deisenroth and Rasmussen, 2011]{deisenroth2011PILCO}
Deisenroth, M. and Rasmussen, C. (2011).
\newblock Pilco: A model-based and data-efficient approach to policy search.
\newblock pages 465--472.

\bibitem[Dempster et~al., 1977]{Dempster77em}
Dempster, A.~P., Laird, N.~M., and Rubin, D.~B. (1977).
\newblock Maximum likelihood from incomplete data via the {EM} algorithm.
\newblock {\em JOURNAL OF THE ROYAL STATISTICAL SOCIETY, SERIES B},
  39(1):1--38.

\bibitem[Dua and Graff, 2017]{dua2017UCI}
Dua, D. and Graff, C. (2017).
\newblock {UCI} machine learning repository.

\bibitem[Ebbehoj et~al., 2021]{Ebbehoj2021TLforHCML}
Ebbehoj, A., Thunbo, M., Andersen, O.~E., Glindtvad, M.~V., and Hulman, A.
  (2021).
\newblock Transfer learning for non-image data in clinical research: a scoping
  review.
\newblock {\em medRxiv}.

\bibitem[Feng et~al., 2020]{feng2020sgbm}
Feng, J., Xu, Y.-X., Jiang, Y., and Zhou, Z.-H. (2020).
\newblock Soft gradient boosting machine.

\bibitem[Flam-Shepherd, 2017]{FlamShepherd2017MappingGP}
Flam-Shepherd, D. (2017).
\newblock Mapping {G}aussian process priors to {B}ayesian neural networks.

\bibitem[Fortuin et~al., 2021]{fortuin2021bnnpriorrevisited}
Fortuin, V., Garriga-Alonso, A., Ober, S.~W., Wenzel, F., Rätsch, G., Turner,
  R.~E., van~der Wilk, M., and Aitchison, L. (2021).
\newblock {B}ayesian neural network priors revisited.

\bibitem[Freund and Schapire, 1997]{FREUND1997Adaboost}
Freund, Y. and Schapire, R.~E. (1997).
\newblock A decision-theoretic generalization of on-line learning and an
  application to boosting.
\newblock {\em Journal of Computer and System Sciences}, 55(1):119--139.

\bibitem[Friedman, 2001]{Friedman2001GBM}
Friedman, J.~H. (2001).
\newblock {Greedy function approximation: A gradient boosting machine.}
\newblock {\em The Annals of Statistics}, 29(5):1189 -- 1232.

\bibitem[Frosst and Hinton, 2017]{frosst1017distilling}
Frosst, N. and Hinton, G. (2017).
\newblock Distilling a neural network into a soft decision tree.

\bibitem[Goldberg et~al., 2001]{Goldberg2001jester}
Goldberg, K., Roeder, T., Gupta, D., and Perkins, C. (2001).
\newblock Eigentaste: A constant time collaborative filtering algorithm.
\newblock {\em Information Retrieval}, 4:133--151.

\bibitem[Gonzalvez et~al., 2019]{bayesMLinFinance}
Gonzalvez, J., Lezmi, E., Roncalli, T., and Xu, J. (2019).
\newblock Financial applications of {G}aussian processes and {B}ayesian
  optimization.

\bibitem[Graf et~al., 2011]{Graf20112DIR}
Graf, F., Kriegel, H.-P., Schubert, M., P{\"o}lsterl, S., and Cavallaro, A.
  (2011).
\newblock {2D} image registration in {CT} images using radial image
  descriptors.
\newblock {\em Medical image computing and computer-assisted intervention :
  MICCAI ... International Conference on Medical Image Computing and
  Computer-Assisted Intervention}, 14 Pt 2:607--14.

\bibitem[Guo et~al., 2017]{guo2017calibration}
Guo, C., Pleiss, G., Sun, Y., and Weinberger, K.~Q. (2017).
\newblock On calibration of modern neural networks.
\newblock In {\em International conference on machine learning}, pages
  1321--1330. PMLR.

\bibitem[He and Hahn, 2020]{he2020stochastictreeensemble}
He, J. and Hahn, P.~R. (2020).
\newblock Stochastic tree ensembles for regularized nonlinear regression.

\bibitem[Hennigan et~al., 2020]{haiku2020github}
Hennigan, T., Cai, T., Norman, T., and Babuschkin, I. (2020).
\newblock {H}aiku: {S}onnet for {JAX}.

\bibitem[Hoang and Wiegratz, 2021]{Hoang2021MachineLM}
Hoang, D. and Wiegratz, K. (2021).
\newblock Machine learning methods in finance: Recent applications and
  prospects.

\bibitem[Hoffman et~al., 2013]{hoffman2013StochasticVI}
Hoffman, M.~D., Blei, D.~M., Wang, C., and Paisley, J. (2013).
\newblock Stochastic variational inference.
\newblock {\em Journal of Machine Learning Research}, 14(40):1303--1347.

\bibitem[Immer et~al., 2020]{immer2020BNNlocalLaplace}
Immer, A., Korzepa, M., and Bauer, M. (2020).
\newblock Improving predictions of {B}ayesian neural nets via local
  linearization.

\bibitem[Irsoy et~al., 2012]{isroy2012softdecisiontree}
Irsoy, O., Yildiz, O., and Alpaydın, E. (2012).
\newblock Soft decision trees.
\newblock pages 1819--1822.

\bibitem[Jordan and Jacobs, 1993]{jordan1993HME}
Jordan, M. and Jacobs, R. (1993).
\newblock Hierarchical mixtures of experts and the {EM} algorithm.
\newblock In {\em Proceedings of 1993 International Conference on Neural
  Networks (IJCNN-93-Nagoya, Japan)}, volume~2, pages 1339--1344 vol.2.

\bibitem[Kingma and Ba, 2014]{kingma2014adam}
Kingma, D.~P. and Ba, J. (2014).
\newblock Adam: A method for stochastic optimization.

\bibitem[Kingma and Welling, 2013]{kingma2013VAE}
Kingma, D.~P. and Welling, M. (2013).
\newblock Auto-encoding variational {B}ayes.

\bibitem[Kirsch et~al., 2019]{kirsch2019BatchBald}
Kirsch, A., van Amersfoort, J., and Gal, Y. (2019).
\newblock Batchbald: Efficient and diverse batch acquisition for deep
  {B}ayesian active learning.

\bibitem[Kompa et~al., 2021]{uncertaintymedicalML}
Kompa, B., Snoek, J., and Beam, A. (2021).
\newblock Second opinion needed: communicating uncertainty in medical machine
  learning.
\newblock {\em npj Digital Medicine}, 4.

\bibitem[Kontschieder et~al., 2015]{kontschieder2015deepneuralforest}
Kontschieder, P., Fiterau, M., Criminisi, A., and Rota~Bulò, S. (2015).
\newblock Deep neural decision forests.
\newblock pages 1467--1475.

\bibitem[Krause et~al., 2006]{krause2006sensorplacementGP}
Krause, A., Guestrin, C., Gupta, A., and Kleinberg, J. (2006).
\newblock Near-optimal sensor placements: Maximizing information while
  minimizing communication cost.
\newblock In {\em 2006 5th International Conference on Information Processing
  in Sensor Networks}, pages 2--10.

\bibitem[Krishnan et~al., 2020]{Krishnan_Subedar_Tickoo_2020}
Krishnan, R., Subedar, M., and Tickoo, O. (2020).
\newblock Specifying weight priors in {B}ayesian deep neural networks with
  empirical {B}ayes.
\newblock {\em Proceedings of the AAAI Conference on Artificial Intelligence},
  34(04):4477--4484.

\bibitem[Kristiadi et~al., 2020]{kristiadi2020bayesian}
Kristiadi, A., Hein, M., and Hennig, P. (2020).
\newblock Being {B}ayesian, even just a bit, fixes overconfidence in {ReLU}
  networks.

\bibitem[Linero and Yang, 2017]{linero2017SBART}
Linero, A.~R. and Yang, Y. (2017).
\newblock {B}ayesian regression tree ensembles that adapt to smoothness and
  sparsity.

\bibitem[Liu et~al., 2021]{liuhal02957135}
Liu, Z., Pavao, A., Xu, Z., Escalera, S., Ferreira, F., Guyon, I., Hong, S.,
  Hutter, F., Ji, R., Jacques~Junior, J. C.~S., Li, G., Lindauer, M., Luo, Z.,
  Madadi, M., Nierhoff, T., Niu, K., Pan, C., Stoll, D., Treguer, S., Wang, J.,
  Wang, P., Wu, C., Xiong, Y., Zela, A., and Zhang, Y. (2021).
\newblock {Winning solutions and post-challenge analyses of the ChaLearn AutoDL
  challenge 2019}.
\newblock {\em {IEEE Transactions on Pattern Analysis and Machine
  Intelligence}}.

\bibitem[Luo et~al., 2021]{Luo2021sdtr}
Luo, H., Cheng, F., Yu, H., and Yi, Y. (2021).
\newblock {SDTR}: Soft decision tree regressor for tabular data.
\newblock {\em IEEE Access}, 9:55999--56011.

\bibitem[Ma and Hern\'{a}ndez-Lobato, 2021]{ma2021FVISPG}
Ma, C. and Hern\'{a}ndez-Lobato, J.~M. (2021).
\newblock Functional variational inference based on stochastic process
  generators.
\newblock In Ranzato, M., Beygelzimer, A., Dauphin, Y., Liang, P., and Vaughan,
  J.~W., editors, {\em Advances in Neural Information Processing Systems},
  volume~34, pages 21795--21807. Curran Associates, Inc.

\bibitem[{Maia} et~al., 2022]{maia2022GPBART}
{Maia}, M., {Murphy}, K., and {Parnell}, A.~C. (2022).
\newblock {{GP-BART}: A novel {B}ayesian additive regression trees approach
  using {G}aussian processes}.
\newblock {\em arXiv e-prints}, page arXiv:2204.02112.

\bibitem[Malinin et~al., 2020]{malinin2021uncertGBM}
Malinin, A., Prokhorenkova, L., and Ustimenko, A. (2020).
\newblock Uncertainty in gradient boosting via ensembles.

\bibitem[Mitros and Namee, 2019]{mitros2019validity}
Mitros, J. and Namee, B.~M. (2019).
\newblock On the validity of {B}ayesian neural networks for uncertainty
  estimation.

\bibitem[Mukhoti et~al., 2021]{Mukhoti2021ddu}
Mukhoti, J., Kirsch, A., van Amersfoort, J., Torr, P. H.~S., and Gal, Y.
  (2021).
\newblock Deep deterministic uncertainty: A simple baseline.

\bibitem[Mustafa and Rahimi Azghadi, 2021]{computers10020024}
Mustafa, A. and Rahimi Azghadi, M. (2021).
\newblock Automated machine learning for healthcare and clinical notes
  analysis.
\newblock {\em Computers}, 10(2).

\bibitem[Niculescu-Mizil and Caruana, 2012]{Niculescu2012calibrationGBM}
Niculescu-Mizil, A. and Caruana, R.~A. (2012).
\newblock Obtaining calibrated probabilities from boosting.

\bibitem[Pratola et~al., 2017]{pratola2017hbart}
Pratola, M., Chipman, H., George, E., and McCulloch, R. (2017).
\newblock Heteroscedastic {BART} using multiplicative regression trees.

\bibitem[Riquelme et~al., 2018]{riquelme2018DeepBayesianBandit}
Riquelme, C., Tucker, G., and Snoek, J. (2018).
\newblock Deep {B}ayesian bandits showdown: An empirical comparison of
  {B}ayesian deep networks for {Thompson} sampling.

\bibitem[Shwartz-Ziv and Armon, 2022]{SHWARTZZIV202284}
Shwartz-Ziv, R. and Armon, A. (2022).
\newblock Tabular data: Deep learning is not all you need.
\newblock {\em Information Fusion}, 81:84--90.

\bibitem[Springenberg et~al., 2016]{springenberg2016BOBNN}
Springenberg, J.~T., Klein, A., Falkner, S., and Hutter, F. (2016).
\newblock {B}ayesian optimization with robust {B}ayesian neural networks.
\newblock In Lee, D., Sugiyama, M., Luxburg, U., Guyon, I., and Garnett, R.,
  editors, {\em Advances in Neural Information Processing Systems}, volume~29.
  Curran Associates, Inc.

\bibitem[Swiatkowski et~al., 2020]{swiatkowski2020ktied}
Swiatkowski, J., Roth, K., Veeling, B.~S., Tran, L., Dillon, J.~V., Snoek, J.,
  Mandt, S., Salimans, T., Jenatton, R., and Nowozin, S. (2020).
\newblock The k-tied normal distribution: A compact parameterization of
  {G}aussian mean field posteriors in {B}ayesian neural networks.

\bibitem[Thompson, 1933]{thompson1933TS}
Thompson, W.~R. (1933).
\newblock {On the likelihood that one unknown probability exceeds another in
  view of the evidence of two samples}.
\newblock {\em Biometrika}, 25(3-4):285--294.

\bibitem[Tomczak et~al., 2020]{tomczak2020lrvinn}
Tomczak, M., Swaroop, S., and Turner, R. (2020).
\newblock Efficient low rank {G}aussian variational inference for neural
  networks.
\newblock In Larochelle, H., Ranzato, M., Hadsell, R., Balcan, M., and Lin, H.,
  editors, {\em Advances in Neural Information Processing Systems}, volume~33,
  pages 4610--4622. Curran Associates, Inc.

\bibitem[Tran et~al., 2015]{tran2015vigp}
Tran, D., Ranganath, R., and Blei, D.~M. (2015).
\newblock The variational {G}aussian process.

\bibitem[Ueda and Ghahramani, 2002]{ueda2002BayesianHME}
Ueda, N. and Ghahramani, Z. (2002).
\newblock {B}ayesian model search for mixture models based on optimizing
  variational bounds.
\newblock {\em Neural Networks}, 15(10):1223--1241.

\bibitem[Ustimenko and Prokhorenkova, 2020]{ustimenko2020sglb}
Ustimenko, A. and Prokhorenkova, L. (2020).
\newblock {SGLB}: Stochastic gradient {Langevin} boosting.

\bibitem[Vladimirova et~al., 2018]{Vladimirova2018understandingpriorsbnn}
Vladimirova, M., Verbeek, J., Mesejo, P., and Arbel, J. (2018).
\newblock Understanding priors in {B}ayesian neural networks at the unit level.

\bibitem[Waterhouse et~al., 1995]{waterhouse1995BayesianHME}
Waterhouse, S., MacKay, D., and Robinson, A. (1995).
\newblock {B}ayesian methods for mixtures of experts.
\newblock In Touretzky, D., Mozer, M., and Hasselmo, M., editors, {\em Advances
  in Neural Information Processing Systems}, volume~8. MIT Press.

\bibitem[Welling and Teh, 2011]{welling2011sgld}
Welling, M. and Teh, Y.~W. (2011).
\newblock {B}ayesian learning via stochastic gradient {Langevin} dynamics.
\newblock In {\em Proceedings of the 28th International Conference on
  International Conference on Machine Learning}, ICML'11, page 681–688,
  Madison, WI, USA. Omnipress.

\end{thebibliography}

\setcounter{section}{0}

\onecolumn
\section{Additional method details}

\subsection{Variational inference}
\label{sec:vi}

We use the variational distribution proposed by \citet{tomczak2020lrvinn}, which is a normal distribution with a low-rank covariance matrix:
\begin{align}
    q(\theta) = \gaussianx{\theta}{\mu}{\text{diag}[\sigma^2] + V V^T} \label{eq:supp_posterior}
\end{align}
where $\mu \in \mathbb{R}^p$ is the mean of the normal distribution over parameters, $\text{diag}[\sigma^2] \in \mathbb{R}^{p\times p}$ is the diagonal component of the covariance with $\sigma \in \mathbb{R}^p$, and $V \in \mathbb{R}^{p \times k}$ is the rank $k$ component of the covariance.
The hyperparameter $k$ determines the rank of the covariance matrix, and hence the expressivity of the posterior.
Choosing $k = p$ allows learning arbitrary covariance matrices, but increases the memory requirements and computational cost.
As a result, in our experiments we typically choose $k < p$. 
We fit the variational soft tree by maximizing the evidence lower bound (ELBO):
\begin{align}
    \text{ELBO} = \expect{\log \prob{y \g x, \theta}}{\theta \sim q\br{\theta}} - \kl{q\br{\theta}}{p\br{\theta}}.
\end{align}
We compute a Monte Carlo estimate of the expectation: 
\begin{align}
\expect{\log \prob{y \g x, \theta}}{\theta \sim q} \approx \frac{1}{m} \sum_{i=1}^m \log \prob{y \g x, \theta^{(i)}},
\end{align}
where $\theta^{(i)} \sim q\br{\theta}$.
For the chosen posterior in \cref{eq:supp_posterior} and an isotropic Gaussian prior $\prob{\theta} = \gaussianx{\theta}{0}{\gamma \eye}$, the KL divergence term is available in a closed form:
\begin{align}
    \kl{q\br{\theta}}{p\br{\theta}} = \frac{1}{2}\left[ \sum_{d=1}^p \left( \frac{\sigma_d^2}{\gamma}-\log\:\sigma_d^2\right) + \frac{1}{\gamma}\sum_{i=1}^k ||v_i||_2^2 - \Delta + \frac{1}{\gamma} ||\mu||_2^2 +p(\log\:\gamma - 1)\right],
\end{align}
where $V = [v_1, \dots v_k] \in \mathbb{R}^{p \times k}$, $\sigma \in \mathbb{R}^p$ and $\Delta = \log \det(\eye_k+V^T \text{diag}[{\sigma^2}]^{-1}V)$.

\section{Experimental details}

\subsection{Regression}

We consider a suite of regression datasets from \citet{malinin2021uncertGBM} to measure the expressivity of our model. 
\cref{tab:dataset_description} summarizes the number of examples and features in each dataset. 
We preprocess the datasets by encoding categorical variables as one-hot vectors and normalizing each feature to have zero mean and unit variance.
We further split the datasets into train (80\%) and test (20\%) data, and perform 10-fold cross validation on the training data. 
We report the mean and the standard deviation of the test log-likelihood and root mean square error (RMSE) across all folds.

\begin{table}[h]
\scshape
\caption{Regression dataset description}
\label{tab:dataset_description}
\resizebox{\linewidth}{!}{
\begin{tabular}{@{}lllllllllll@{}}
\toprule
Dataset & Boston & Naval  & Power & Protein & Yacht & Song & Concrete & Energy & Kin8nm & Wine \\ \midrule
Number samples & 506 & 11\,934 & 9\,568 & 45\,730 & 308 & 514\,345 & 1\,030 & 768 & 8\,192 & 1\,599 \\
Number features & 13 & 16 & 4 & 9 & 6 & 90 & 8 & 8 & 8 & 11\\ \bottomrule
\end{tabular}
}
\end{table}

\subsection{OOD detection}

Following the setup from \citet{malinin2021uncertGBM}, we consider the test data to be the in-distribution (ID) data, and create a dataset of OOD data by taking a subset of the \textit{song} dataset of the same size. 
We select a subset of features from the \textit{song} dataset in order to match the feature dimension of the ID dataset. 
The only exception is the \textit{song} dataset itself, for which we use the \textit{Relative location of CT slices on axial axis} dataset from \citet{Graf20112DIR} as OOD data.
In both settings, categorical features are encoded as one-hot vectors, and all features are normalized to have zero mean and unit variance. 
We use a one-node decision tree to find the splitting threshold for the epistemic uncertainty that would separate ID and ODD data.

\subsection{Bandit}
\label{app:bandit_datasets}

We consider the following datasets for our bandit experiments:

\paragraph{Financial} The financial bandit \citep{riquelme2018DeepBayesianBandit} was created by pulling the stock prices of $d=21$ publicly traded companies in NYSE and Nasdaq, for the last 14 years for a total of $n=3713$ samples. For each day, the context was the price difference between the beginning and end of the session for each stock. The arms are synthetically created to be a linear combination of the contexts, representing $k=8$ different portfolios. 

\paragraph{Jester} This bandit problem is taken from \citet{riquelme2018DeepBayesianBandit}. The Jester dataset \citep{Goldberg2001jester} provides continuous ratings in $[-10, 10]$ for 100 jokes from 73\,421 users. The authors find a complete subset of $n=19\,181$ users rating all 40 jokes. We take $d=32$ of the ratings as the context of the user, and $k=8$ as the arms. The agent recommends one joke, and obtains the reward corresponding to the rating of the user for the selected joke. 

\paragraph{Exploration} The exploration bandit tests the extent to which the models explore the different arms. The reward function of each arm is sparse and requires maintaining high uncertainty to discover the context values for which the reward is high. The reward given input $x$ and selected arm $a$ is given by: 
\begin{align}
    r(x, a) = \sigma(\beta(x+\alpha-\text{offset}[a]))-
    \sigma(\beta(x-\alpha-\text{offset}[a])) + \epsilon
\end{align}
where $\sigma(x) = 1 \mathbin{/}(1+e^{-x})$ is the sigmoid function, $\alpha, \beta \in \mathbb{R}$ are constants, $\epsilon \sim N(0, \delta^2)$, and $\text{offset}[a]$ is a function mapping each arm $a$ to a particular offset. Intuitively, this reward is composed of a smooth unit valued “bump” per arm and is otherwise 0. The positions of the “bumps” across arms are disjoint and barely overlap. Furthermore, the context is sampled uniformly at random in the interval $[-1, 1]$. We run this bandit for 20\,000 steps. 
  
\paragraph{Mushroom} This bandit problem is taken from \citet{riquelme2018DeepBayesianBandit}. The Mushroom Dataset \citep{dua2017UCI} contains $n=7\,310$ samples, 22 attributes per mushroom, and two classes: poisonous and safe. The agent must decide whether to eat a given mushroom or not. Eating a safe mushroom provides reward of 5. Eating a poisonous mushroom delivers reward 5 with probability 1/2 and reward -35 otherwise. If the agent does not eat a mushroom, then the reward is 0. 

\subsection{Implementation}

We implement the variational soft decision tree and variational soft GBM using the Jax \citep{jax2018github} and DM-Haiku \citep{haiku2020github} Python libraries.
We fit our models using the Adam \citep{kingma2014adam} optimizer. 
Hyperparameter optimization was conducted using Amazon SageMaker's\footnote{\url{https://aws.amazon.com/sagemaker/}} automatic model tuning that uses Bayesian optimisation to explore the hyperparameter space.
We define the set of hyperparameters that the Bayesian optimisation needs to explore, as well as the optimization objective. 
The service then runs a series of model tuning jobs with different hyperparameters to find the set of values which minimize the objective.  
We use the same hyperparameter ranges as \citep{he2020stochastictreeensemble} and \citep{ustimenko2020sglb}, summarized in \cref{tab:hp_tuning_ranges}.

\begin{table}[h]
\scshape
\caption{Hyperparameter tuning ranges.}
\label{tab:hp_tuning_ranges}
\resizebox{\linewidth}{!}{
\begin{tabular}{@{}ll | ll | ll | ll | ll | ll@{}}
\toprule
\multicolumn{2}{c}{VSGBM} &  \multicolumn{2}{c}{XBART} & \multicolumn{2}{c}{XGBOOST} & \multicolumn{2}{c}{Deep ensemble} & \multicolumn{2}{c}{VST} & \multicolumn{2}{c}{SGLB}  \\
\midrule
Learning rate & $[0.0001, 0.001]$ & s & $[1, 300]$ & Learning rate & $[0.001, 0.1]$ & Learning rate & $[0.0001, 0.001]$ & Learning rate & $[0.0001, 0.001]$ & Learning rate & $\set{0.001, 0.01, 0.1}$ \\
Prior scale & $[0.01, 2.5]$ & Tau & $[1, 300]$ & Tree depth & $\set{1, \dots, 6}$ & Num. layers & $\set{2, \dots, 5}$ & Prior scale & $[0.01, 2.5]$ & Tree depth & $\set{3, \dots, 6}$ \\
Max. depth & $\set{1, \dots, 5}$ & Beta & $[0.75, 2]$ & Lambda & $[0.001, 1]$ & Hidden dim. & $\set{10, \dots, 50}$ & Max. depth & $\set{1, \dots, 5}$ & Num. learners & $\set{1, \dots, 1000}$\\
Beta & $[1, 50]$ & Num. learners & $[1, 100]$ & Alpha & $[0.001, 1]$ & Dropout rate & $[0.1, 0.4]$ & Beta & $[1, 50]$ &   &  \\
Num. learners & $\set{1, \dots, 10}$ & Burn-in & $\set{15}$ & Num. learners & $\set{1, \dots, 1000}$ & Prior scale & $[0.001, 2.5]$ &   &   &   &   \\  
Prior sigma alpha & $[2.5, 5]$ & Alpha & $\set{0.95}$ &   &   &   &   &   &   &   &    \\
Prior sigma beta & $[0.1, 5]$ & Kappa & $\set{3}$ &   &   &   &   &   &   &   &    \\
\bottomrule
\end{tabular}
}
\end{table}

\section{Additional experimental results}

\subsection{Regression}

In this section we present additional regression results reporting the root mean square error (RMSE) of evaluated methods across the considered baselines, see \cref{tab:vsdt_vsgbm_vs_baselines_rmse}.
We find that the VST performs well in terms of test RMSE across datasets: our model outperforms the baselines on 3/10 datasets and matching the RMSE of the top baselines on 3/10.
This further confirms that soft decision trees perform well on tabular data \citep{Luo2021sdtr}.

We observe that VSGBMs are generally weaker in terms of RMSE compared to XGBoost. 
XGBoost yields a smaller RMSE on all but three datasets, where VSGBM performs better on one and ties on two. 
This result is not surprising as XGBoost is the gold standard for fitting tabular data. 
Compared to other baselines, VSGBM systematically outperforms deep ensembles, and performs favorably compared to SGLB and XBART. 
Compared to a single variational soft tree, VSGBMs perform better in terms of RMSE on all datasets but two where they ties.

\begin{table*}
\footnotesize
\scshape
\caption{Test RMSE of evaluated methods on regression datasets.
We find that variational soft decision trees (\textsc{VST}) compares well with baselines in terms of test RMSE, out-performing them on 3/10 datasets and matching the RMSE of the best baselines on three. Variational soft GBM (\textsc{VSGBM}) performs competitively in terms of RMSE compared to SGLB, deep ensemble and XBART, but is generally weaker than XGBoost. This is unsurprising as XGBoost is the state-of-the-art for fitting tabular data.
}
\label{tab:vsdt_vsgbm_vs_baselines_rmse}
\resizebox{\linewidth}{!}{
\begin{tabular}{@{}lcccccccccc@{}}
\toprule
Model   &   Power   & Protein & Boston  &   Naval   &   Yacht   &   Song    &   Concrete    & Energy &   Kin8nm  & Wine  \\ \midrule
Deep ensemble   & 0.986 $\pm$ 0.001   & 1.003 $\pm$ 0.001 & 0.870 $\pm$ 0.004 & 0.993 $\pm$ 0.001 & 0.685 $\pm$ 0.002 & 0.999 $\pm$ 0.001   & 0.972 $\pm$ 0.001   & 1.007 $\pm$ 0.009   & 0.998 $\pm$ 0.001   & 1.013 $\pm$ 0.011\\
\midrule
XBART (1 tree)  & 0.243 $\pm$ 0.001 & \textbf{0.693 $\pm$ 0.003}    & \textbf{0.356 $\pm$ 0.026}  & \textbf{0.339 $\pm$ 0.125}    & \textbf{0.052 $\pm$ 0.003}    & 0.680 $\pm$ 0.007   & \textbf{0.474 $\pm$ 0.046}    & 0.406 $\pm$ 0.267 & 0.703 $\pm$ 0.013 & 0.815 $\pm$ 0.006\\
SGLB (1 tree)  & 0.280 $\pm$ 0.006  & 0.869 $\pm$ 0.002 & 0.483 $\pm$ 0.025 &  0.830 $\pm$ 0.007 & 0.405 $\pm$ 0.007 & 0.958 $\pm$ 0.007 & 0.541 $\pm$ 0.010 & 0.193 $\pm$ 0.003   & 0.740 $\pm$ 0.002 & 0.827 $\pm$ 0.004\\
XGBoost (1 tree)    & \textbf{0.242 $\pm$ 0.003} & 0.828 $\pm$ 0.001 & \textbf{0.397 $\pm$ 0.077}    & 0.703 $\pm$ 0.001  & \textbf{0.056 $\pm$ 0.006}    & 0.795 $\pm$ 0.003 & \textbf{0.454 $\pm$ 0.021}  & \textbf{0.059 $\pm$ 0.005}    & 0.776 $\pm$ 0.010 & 0.854 $\pm$ 0.020  \\
VST (const.)   & \textbf{0.239 $\pm$ 0.001}   & 0.827 $\pm$ 0.008 & 0.538 $\pm$ 0.004    & \textbf{0.359 $\pm$ 0.033}  & 0.477 $\pm$ 0.027 & 0.710 $\pm$ 0.010 & 0.592 $\pm$ 0.005   & 0.483 $\pm$ 0.099 & \textbf{0.550 $\pm$ 0.010}    & \textbf{0.788 $\pm$ 0.009} \\
VST (linear)  & 0.242 $\pm$ 0.001 & 0.798 $\pm$ 0.011 & 0.469 $\pm$ 0.003 & \textbf{0.376 $\pm$ 0.002}   & 0.500 $\pm$ 0.004 & \textbf{0.633 $\pm$ 0.006}    & \textbf{0.423 $\pm$ 0.066}    & 0.281 $\pm$ 0.044  & \textbf{0.530 $\pm$ 0.018}    & \textbf{0.781 $\pm$ 0.008}\\   
\midrule
XBART   & 0.233 $\pm$ 0.001 & 0.692 $\pm$ 0.003 & 0.303 $\pm$ 0.022 & 0.152 $\pm$ 0.045 & 0.051 $\pm$ 0.003   & 0.576 $\pm$ 0.003 & 0.355 $\pm$ 0.012 & \textbf{0.292 $\pm$ 0.295} & 0.590 $\pm$ 0.007 & 0.815 $\pm$ 0.008 \\
SGLB    & 0.453 $\pm$ 0.001   & 0.822 $\pm$ 0.001   & 0.583 $\pm$ 0.004 & 0.649 $\pm$ 0.002   & 0.300 $\pm$ 0.004    & 0.834 $\pm$ 0.001   & 0.593 $\pm$ 0.002   & 0.318 $\pm$ 0.002 & 0.778 $\pm$ 0.001   & 0.857 $\pm$ 0.002  \\ 
XGBoost & \textbf{0.173 $\pm$ 0.003}    & \textbf{0.614 $\pm$ 0.004}    & \textbf{0.261 $\pm$ 0.012}    & \textbf{0.057 $\pm$ 0.001}    & \textbf{0.041 $\pm$ 0.006}  & \textbf{0.450 $\pm$ 0.001}    & \textbf{0.285 $\pm$ 0.014}    & \textbf{0.050 $\pm$ 0.001}    & 0.479 $\pm$ 0.019 & \textbf{0.735 $\pm$ 0.014} \\
VSGBM (const.)  & 0.238 $\pm$ 0.001   & 0.652 $\pm$ 0.006 & 0.363 $\pm$ 0.025 & \textbf{0.111 $\pm$ 0.060} & \textbf{0.142 $\pm$ 0.107}    & 0.563 $\pm$ 0.004 & 0.347 $\pm$ 0.016   & 0.227 $\pm$ 0.015 & 0.348 $\pm$ 0.009 & 0.781 $\pm$ 0.008 \\
VSGBM (linear)  & 0.242 $\pm$ 0.001 & 0.663 $\pm$ 0.012 & 0.297 $\pm$ 0.009 & \textbf{0.074 $\pm$ 0.018}     & 0.134 $\pm$ 0.005   & 0.483 $\pm$ 0.007 & 0.341 $\pm$ 0.014   & 0.183 $\pm$ 0.037 & \textbf{0.316 $\pm$ 0.010}    & 0.769 $\pm$ 0.006\\
\bottomrule
\end{tabular}
}
\end{table*} 

\subsection{Uncertainty visualization}
\label{app:uncert_viz}

\begin{figure}
    \centering
    \vspace{-1em}
    \resizebox{\linewidth}{!}{
    \begin{tabular}{ccc}
    \subfloat[VST]{\includegraphics[width=0.3\linewidth]{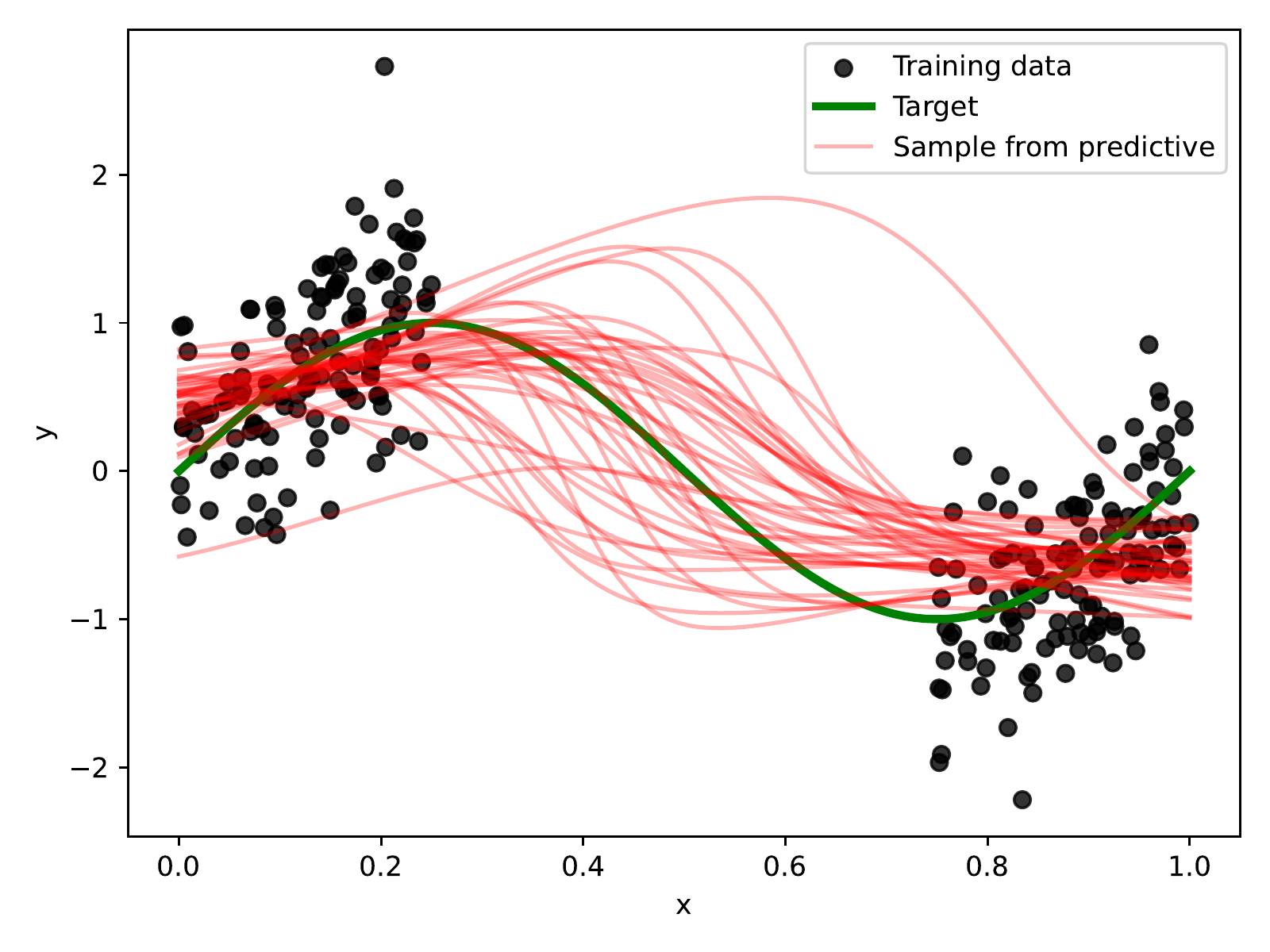}}&
    \subfloat[XBART]{\includegraphics[width=0.3\linewidth]{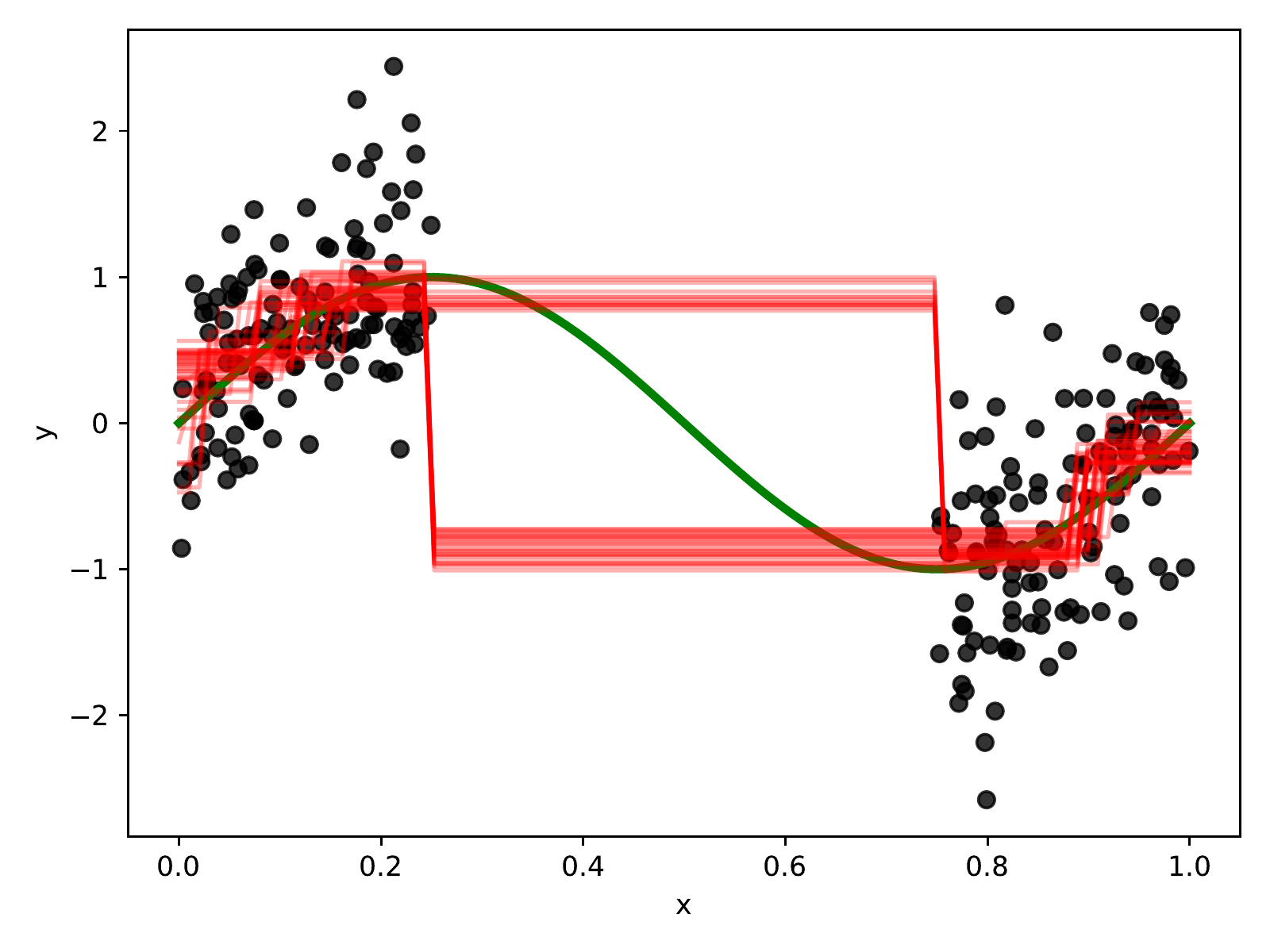}} &
    \subfloat[SGLB]{\includegraphics[width=0.3\linewidth]{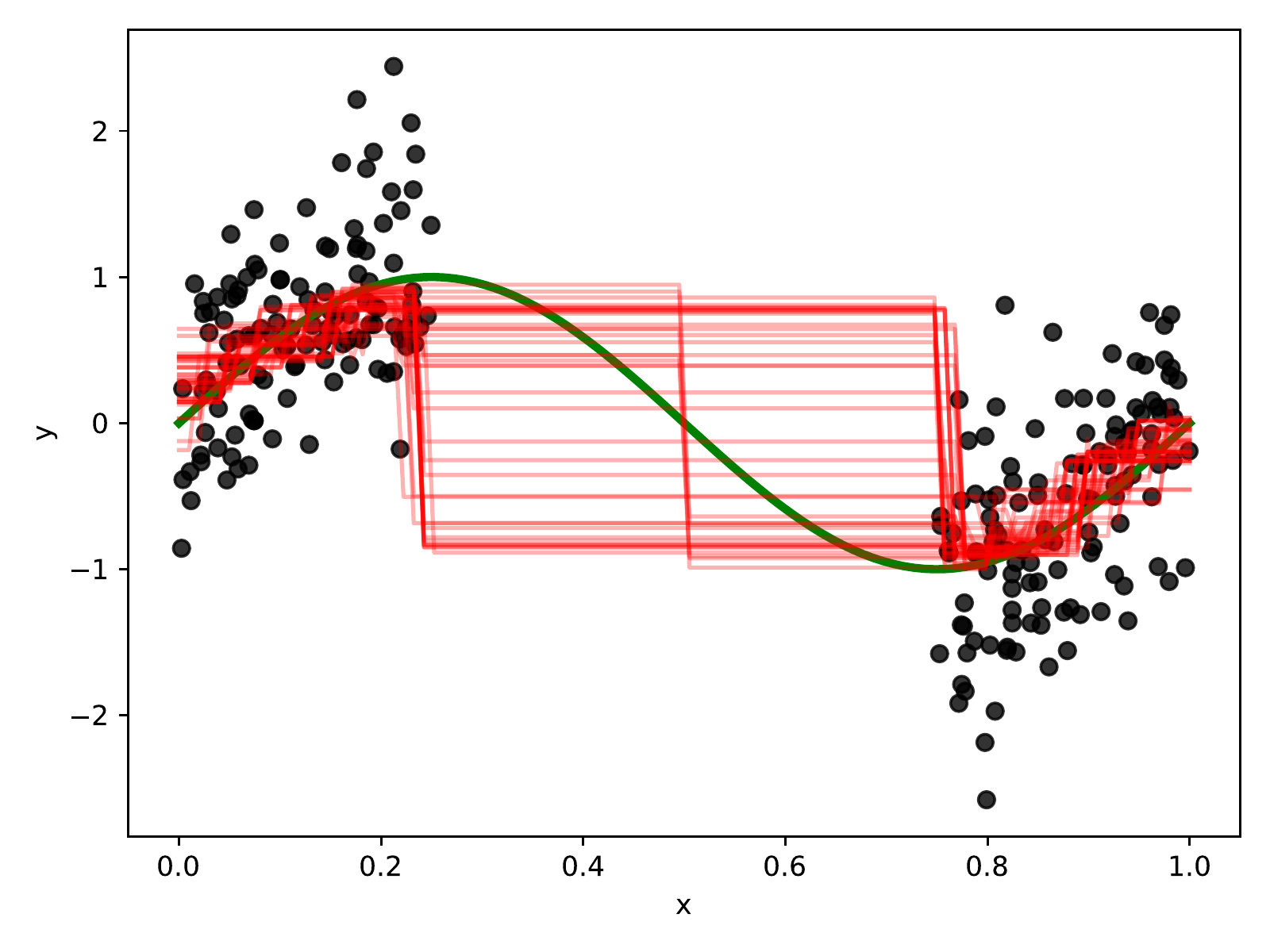}}
    \end{tabular}
    }
    \caption{In-between uncertainty. Variational soft trees (\textsc{VST}) show high uncertainty between the two blobs of data while closely agreeing with the true mean inside of the data support. The plot for the \textsc{VST} was generated with the linear leaf model.}
    \label{fig:app_inbetween_uncertainty}
\end{figure}

In order to demonstrate the quality of the predictive uncertainty of our model, we present additional plots of functions sampled from the VST, comparing these to samples from XBART and SGLB. 
See \cref{fig:tail_uncertainty} and \cref{fig:app_inbetween_uncertainty}. 
We find that variational soft decision trees model the uncertainty outside of the support of the data better than XBART and SGLB: functions agree with the true mean within the support of the data, but diverge outside of it.

\subsection{Regression visualization}

\begin{figure}
    \centering
    \resizebox{\linewidth}{!}{
    \begin{tabular}{ccc}
    \subfloat[VST (RMSE: 0.2121)]{\includegraphics[width=0.3\linewidth]{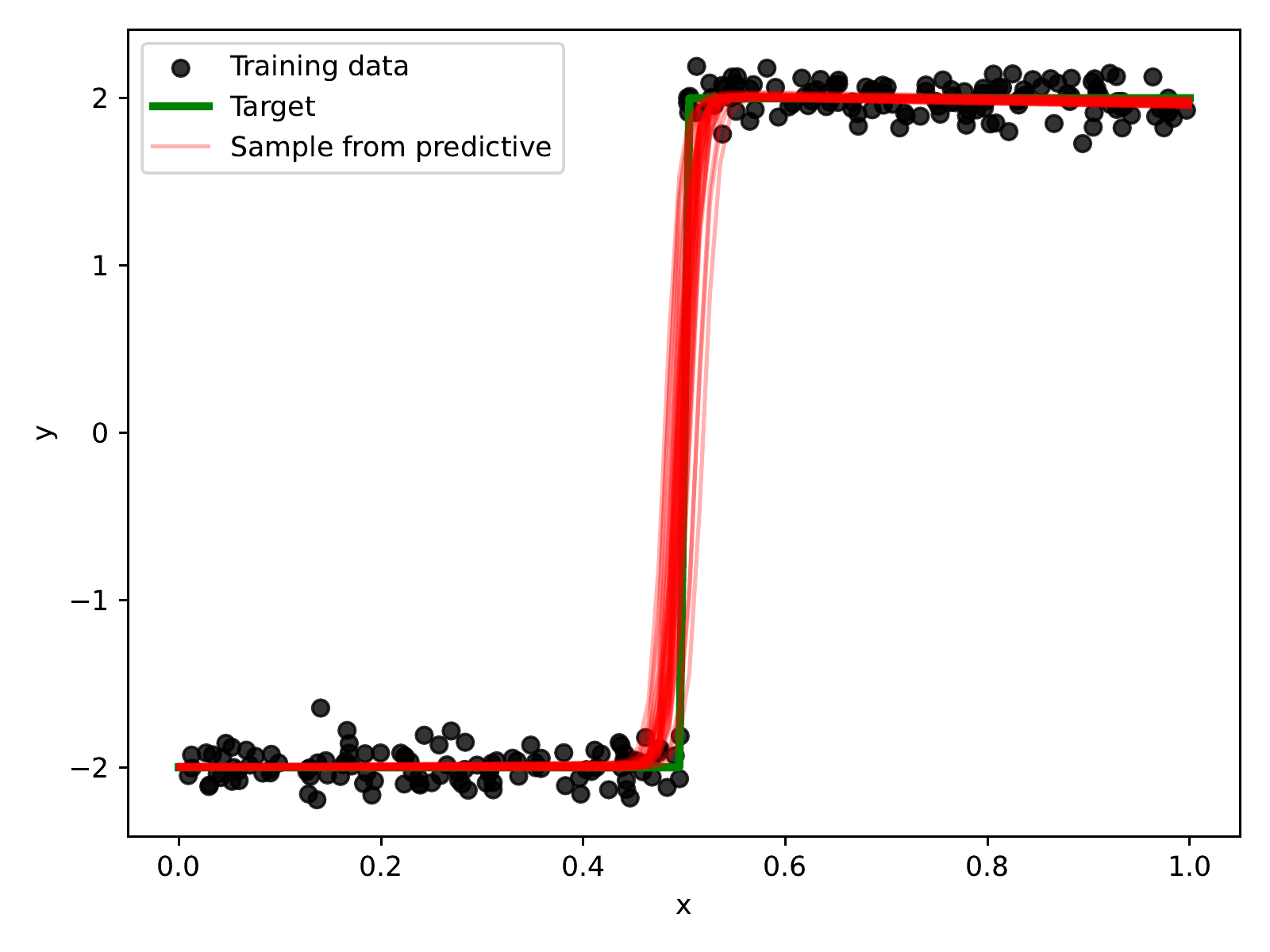}} &
    \subfloat[XBART (RMSE: 0.0982)]{\includegraphics[width=0.3\linewidth]{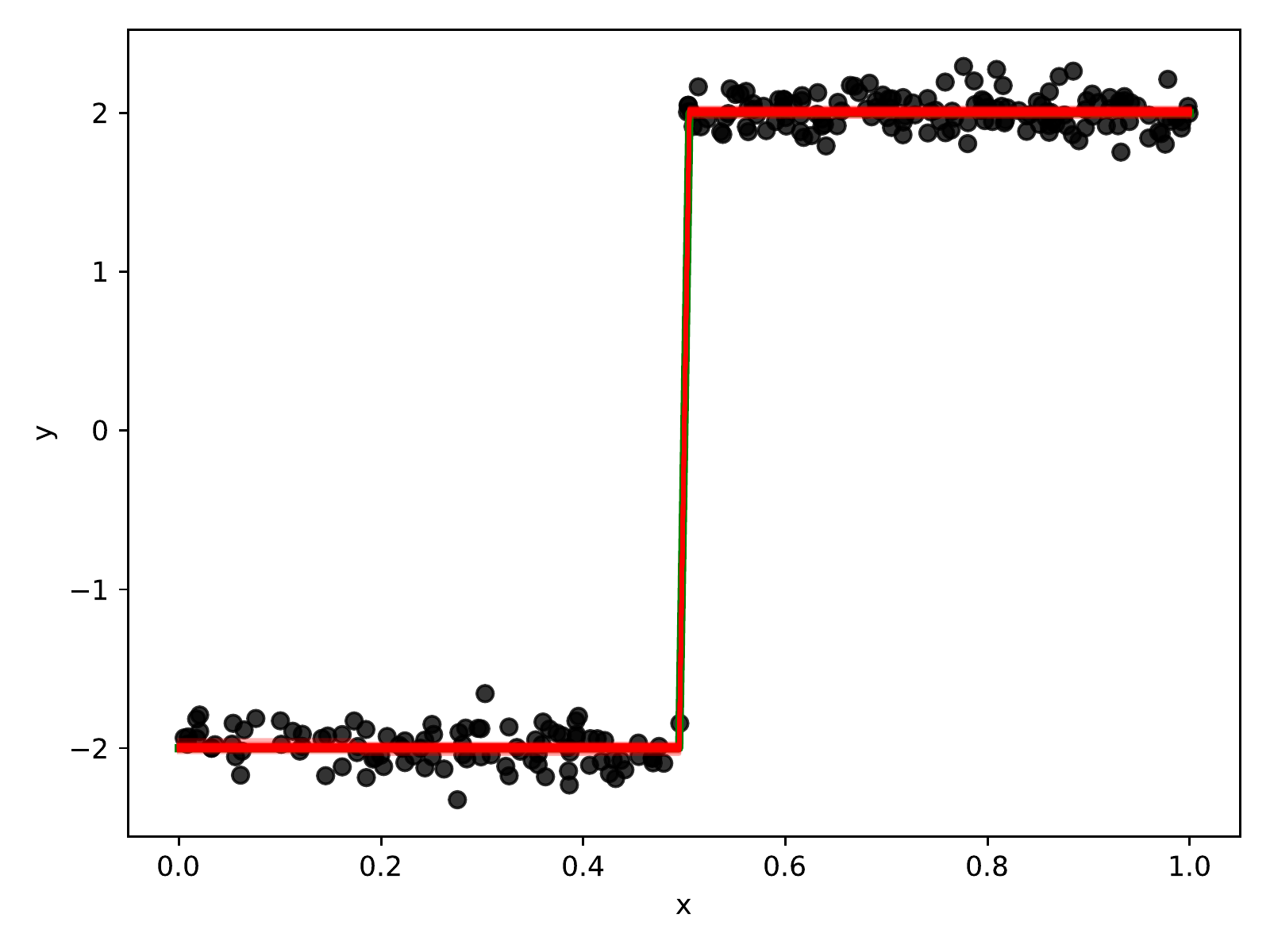}} &
    \subfloat[SGLB (RMSE: 0.3014)]{\includegraphics[width=0.3\linewidth]{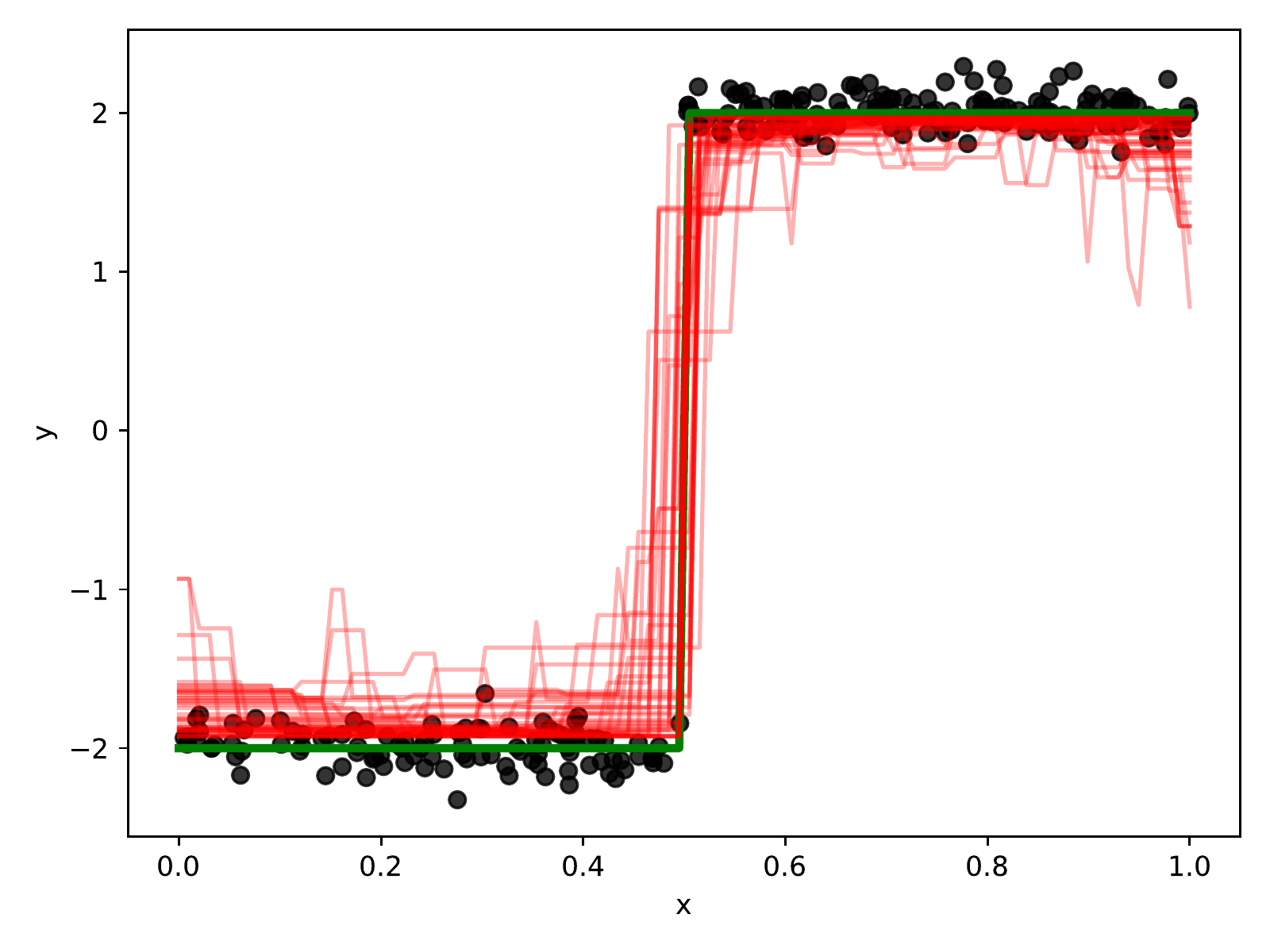}}
    \end{tabular}
    }
    \caption{Step function. The variational soft decision tree (\textsc{VST}) can fit this non-smooth function. The plot for the \textsc{VST} was generated with the linear leaf model.}
    \label{fig:app_non_smooth_function}
\end{figure}

\begin{figure}
    \centering
    \resizebox{0.8\linewidth}{!}{
    \begin{tabular}{cc}
    \subfloat[VST (RMSE: 0.2346)]{\includegraphics[width=0.5\linewidth]{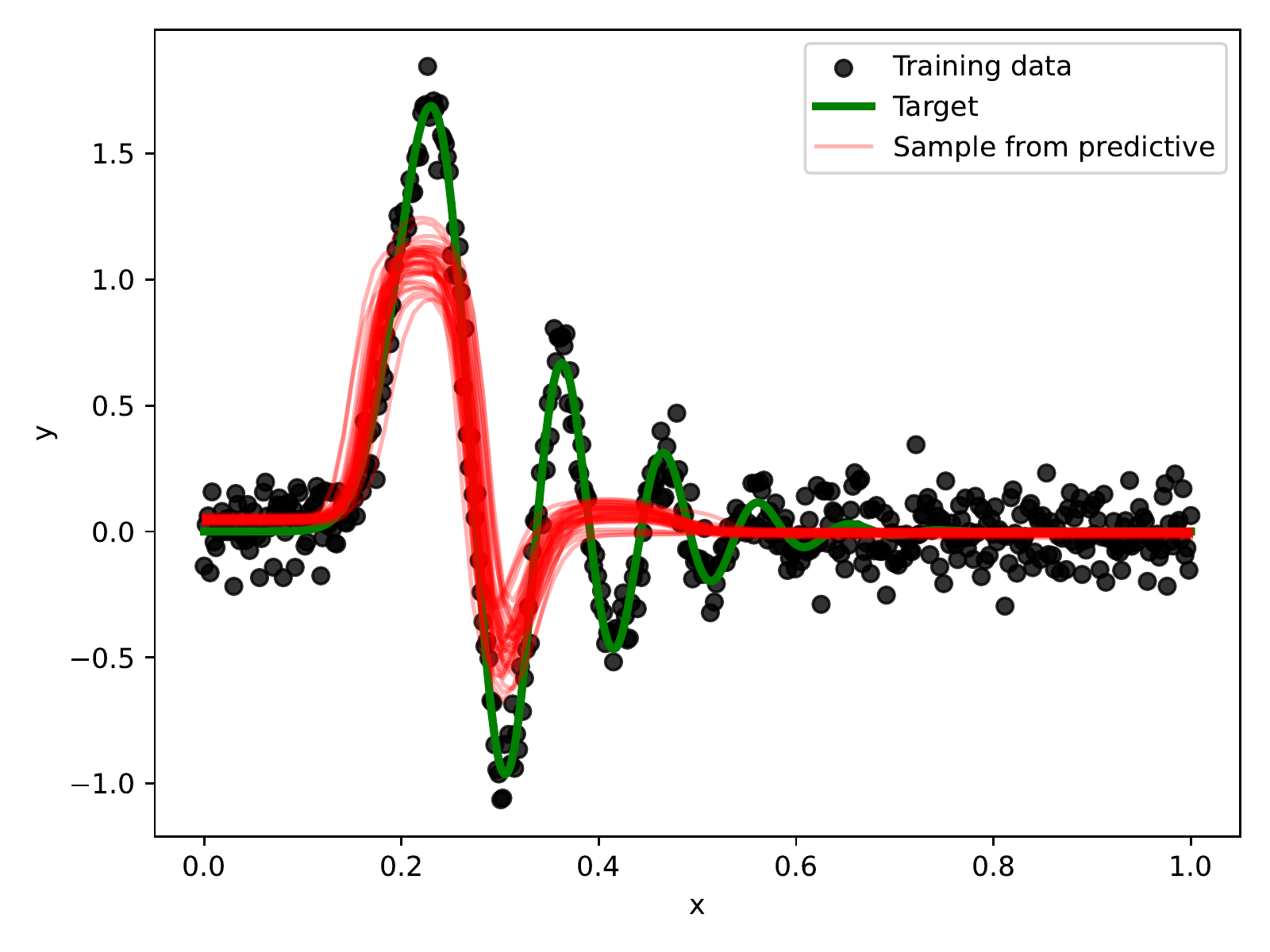}} &
    \subfloat[VSGBM (RMSE: 0.2534)]{\includegraphics[width=0.5\linewidth]{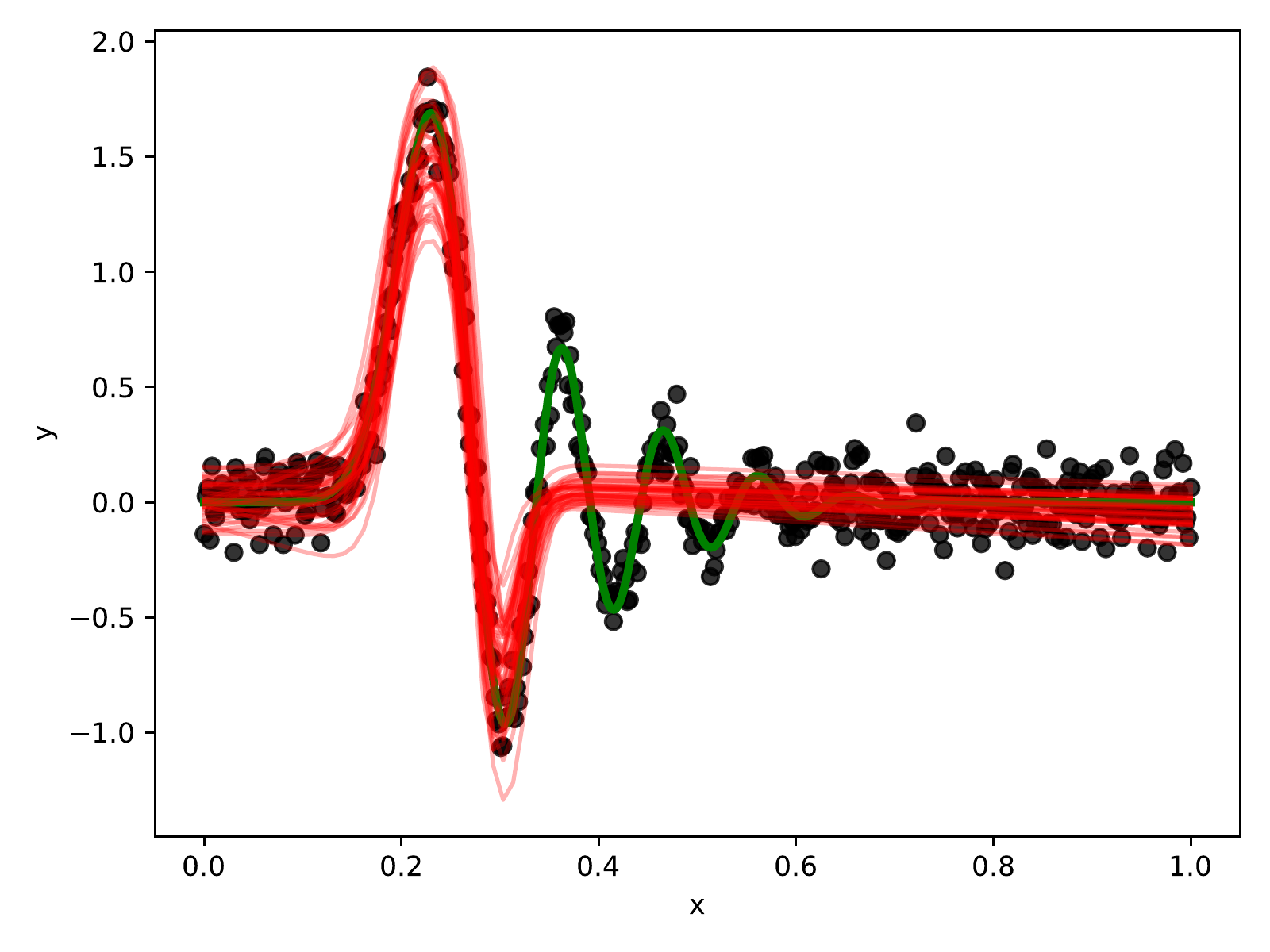}} \\
    \subfloat[XBART (RMSE: 0.1186)]{\includegraphics[width=0.5\linewidth]{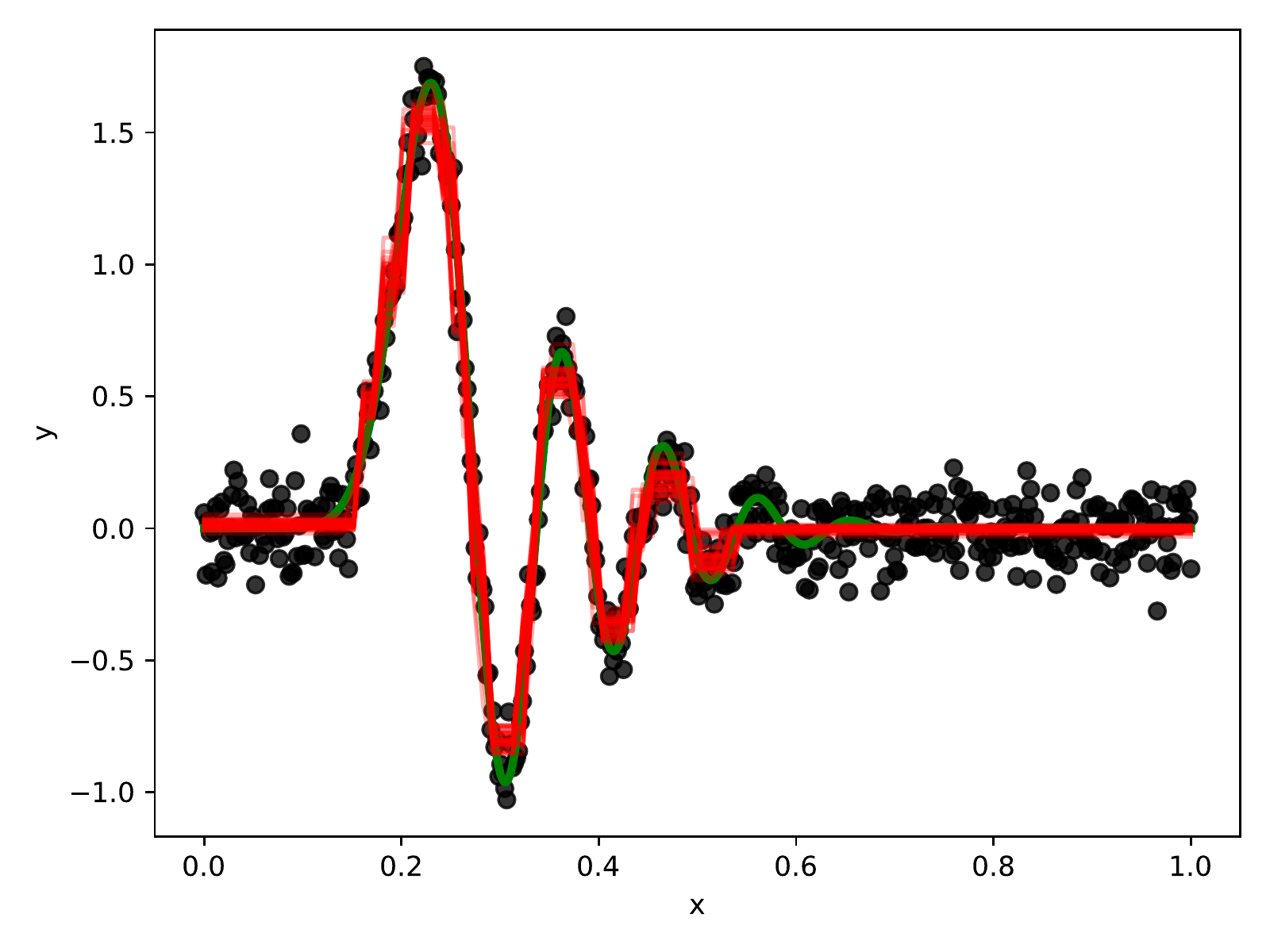}} &
    \subfloat[SGLB (RMSE: 0.1503)]{\includegraphics[width=0.5\linewidth]{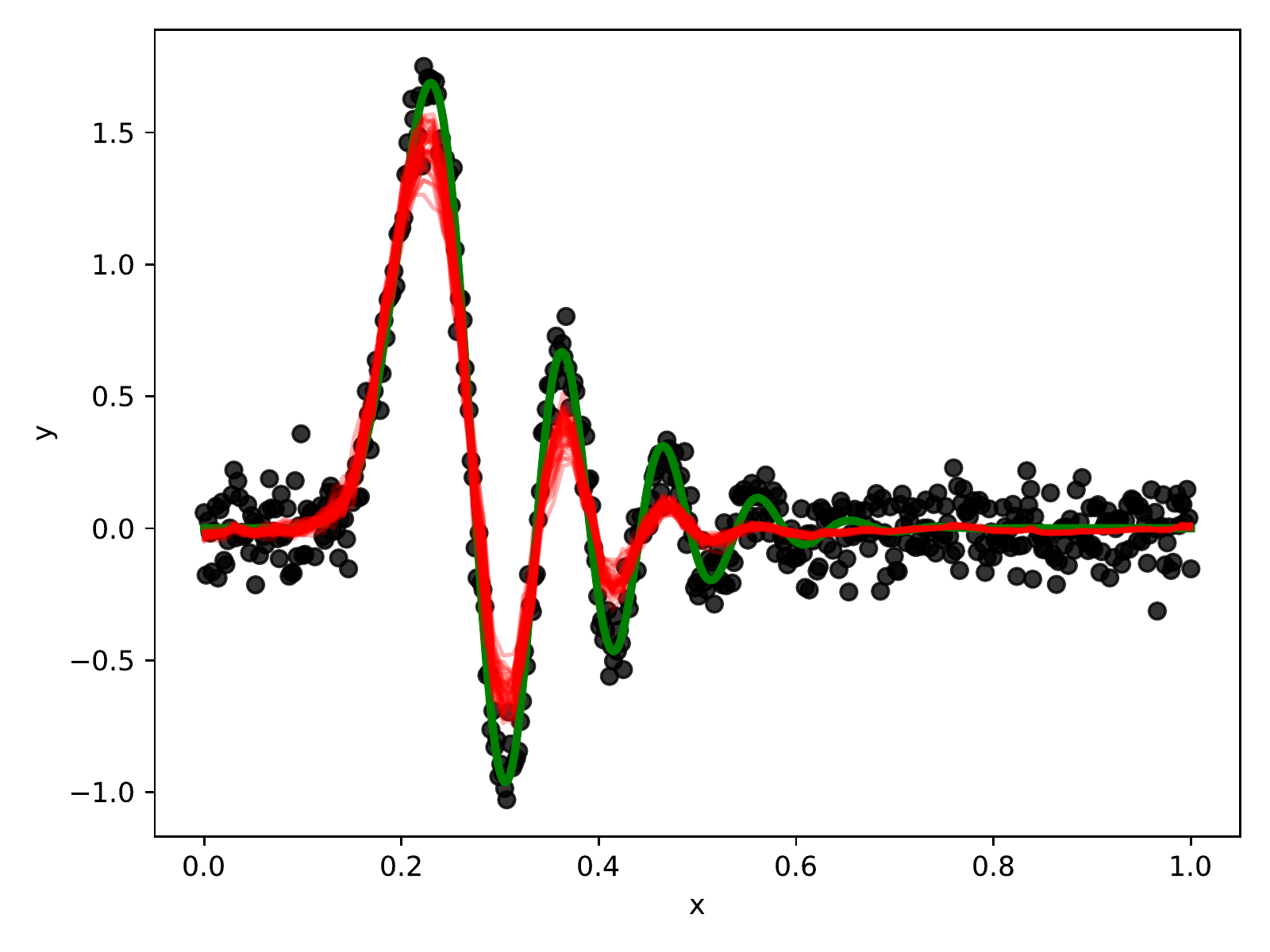}}
    \end{tabular}
    }
    \caption{Daubechies wavelet. The variational soft tree (\textsc{VST}) and variational soft GBM (\textsc{VSGBM}) underfit this complex function.}
    \label{fig:app_daubechies_wavelet}
\end{figure}

To test the flexiblity of the VST in the regression setting, we fit our model to a step function (\cref{fig:app_non_smooth_function}) and a Daubechies wavelet (\cref{fig:app_daubechies_wavelet}) as done in \citep{linero2017SBART}, and plot functions sampled from our model. 
We see that the VST captures the non-smooth step function as well as the hard decision trees used in XBART and SGLB. 
On the Daubechies wavelet, however, we find that a VST of a reasonable depth (5) underfits. 
VSGBM visually improves the fit over a VST, but still underfits the data compared to SGLB and XBART.
We hypothesize that this effect is due to the choice of zero mean Gaussian prior on the node weights $w_n$ in VSTs and VSGBMs, which encourages smooth functions. 
This phenomenon is particularly visible when fitting the Daubechies wavelet, a complex function that fluctuates at different frequencies and amplitudes. 
As expected, increasing the amount of training data reduces underfitting.

\end{document}